%% file: accv2020cameraready.tex
\documentclass[runningheads]{llncs}
\usepackage{graphicx}
\usepackage{amsmath,amssymb} 
\usepackage{color}
\usepackage{mypackages}

\def\myalgonamefirst{TRAS\@\xspace}
\def\myalgonamesecond{TRAST\@\xspace}
\def\myalgonamethird{TRASFUST\@\xspace}

\graphicspath{{./images/}{./figures/}}
\def\reals{\mathbb{R}}

\def\images{\mathcal{I}}

\def\dataset{\mathcal{D}}
\def\video{\mathcal{V}}
\def\episode{E}
\def\mdp{M}
\def\frame{F}

\def\bbox{b}
\def\bboxgt{g}
\def\state{s}
\def\states{\mathcal{S}}
\def\action{a}
\def\actions{\mathcal{A}}
\def\reward{r}
\def\statevalue{v}
\def\counter{C}
\def\contextfactor{c}

\def\tkcf{\texttt{KCF}}
\def\tmdnet{\texttt{MDNet}}
\def\teco{\texttt{ECO}}
\def\tsiamrpn{\texttt{SiamRPN}}
\def\tatom{\texttt{ATOM}}
\def\tdimp{\texttt{DiMP}}

\def\tk{\teachers_{\texttt{K}}}
\def\tm{\teachers_{\texttt{M}}}
\def\te{\teachers_{\texttt{E}}}
\def\ts{\teachers_{\texttt{S}}}
\def\tp{\teachers_{\texttt{P}}}
\def\ta{\teachers_{\texttt{A}}}
\def\td{\teachers_{\texttt{D}}}

\def\IoU{\text{IoU}}

\def\dfiltcoeff{\beta}

\def\loss{\mathcal{L}}

\def\distillationloss{\loss_{\mathrm{dist}}}
\def\policyloss{\loss_{\pi}}
\def\valueloss{\loss_{\statevalue}}

\def\teacher{\mathbf{t}}
\def\teachers{\mathbf{T}}
\def\student{\mathbf{s}}

\def\weights{\theta}

\newcommand*{\smallcup}{\mathbin{\scalebox{0.5}{\ensuremath{\cup}}}}%

\begin{document}
\pagestyle{headings}
\mainmatter

\def\ACCV20SubNumber{141}  

\title{Tracking-by-Trackers \\with a Distilled and Reinforced Model} 
\titlerunning{Tracking-by-Trackers with a Distilled and Reinforced Model}
%
\author{Matteo Dunnhofer \orcidID{0000-0002-1672-667X},
Niki Martinel \orcidID{0000-0002-6962-8643}, \\
Christian Micheloni \orcidID{0000-0003-4503-7483} \\
\footnotesize{\email{\{matteo.dunnhofer, niki.martinel, christian.micheloni\}@uniud.it}}
}


%
\authorrunning{M. Dunnhofer et al.}
%
\institute{Machine Learning and Perception Lab, University of Udine, Italy}

\maketitle

\begin{abstract}
\input{abstract.tex}
\end{abstract}

\input{1introduction.tex}

\input{2relatedwork.tex}

\input{3methods.tex}

\input{4experimental.tex}

\input{5results.tex}

\input{6conclusions.tex}

\subsubsection{Acknowledgements.} This work is supported by the ACHIEVE-ITN project.

\bibliographystyle{splncs}
\bibliography{egbib}

\clearpage
\input{supplementary.tex}

\end{document}

%% file: abstract.tex
Visual object tracking was generally tackled by reasoning independently on fast processing algorithms, accurate online adaptation methods, and fusion of trackers. 
In this paper, we unify such goals by proposing a novel tracking methodology that takes advantage of other visual trackers, offline and online. 
A compact student model is trained via the marriage of knowledge distillation and reinforcement learning. The first allows to transfer and compress tracking knowledge of other trackers. The second enables the learning of  evaluation measures which are then exploited online. After learning, the student can be ultimately used to build (i) a very fast single-shot tracker, (ii) a tracker with a simple and effective online adaptation mechanism, (iii) a tracker that performs fusion of other trackers.
Extensive validation shows that the proposed algorithms compete with real-time state-of-the-art trackers.

%% file: 1introduction.tex
\section{Introduction}
\label{sec:intro}
Visual object tracking corresponds to the persistent recognition and localization --by means of bounding boxes-- of a target object in consecutive video frames.
This problem comes with several different challenges including object occlusion and fast motion, light changes, and motion blur.
Additionally, real-time constraints are often posed by the many practical applications, such as video surveillance, behavior understanding, autonomous driving, and robotics.

In the past, the community has proposed solutions emphasizing different aspects of the problem. Processing speed was pursued by algorithms like correlation filters \cite{Bolme2010,KCF,DSST,Staple,Lukezic2018} or offline methods such as siamese convolutional neural networks (CNNs) \cite{GOTURN,RE3,SiamFC,SiamRPN,SiamRPNpp,DaSiam,Zhang2019}. 
Improved performance was attained by online target adaptation methods \cite{MDNet,RealTimeMDNet,ECO,ATOM,DiMP}.
Higher tracking accuracy and robustness were achieved by methods built on top of other trackers \cite{MEEM,Yoon2012,Wang2014,Bailer2014,HMMTxD}. All these characteristics belong to an optimal tracker but they were studied one independently from the other. The community currently lacks a general framework to tackle them jointly. In this view, a single model should be able to (i) track an object in a fast way, (ii) implement simple and effective online adaptation mechanisms, (iii) apply decision-making strategies to select tracker outputs.

It is a matter of fact that a large number of tracking algorithms has been produced so far, with different principles exploited. 
Preliminary solutions were based on mean shift algorithms \cite{Comanciu2000}, key-point \cite{Matrioska} or part-based methods \cite{LGT,OGT}, or SVM learning \cite{Struck}.
Later, correlation filters gained popularity thanks to their fast processing times \cite{Bolme2010,KCF,DSST,Staple,Lukezic2018}. 
Since more recently, CNNs have been exploited to extract efficient image features. This kind of representation has been included in deep regression networks \cite{GOTURN,RE3}, online tracking-by-detection methods \cite{MDNet,RealTimeMDNet}, solutions that treat visual tracking as a reinforcement learning (RL) problem \cite{Yun2017,Supancic2017,Choi2017,Ren2018,Chen2018,Dunnhofer2019}, CNN-based discriminative correlation filters \cite{CCOT,ECO,ATOM,DiMP}, and in siamese CNNs \cite{SiamFC,SiamRPN,SiamRPNpp,Zhang2019,SiamMask,Dunnhofer2020MedIA}. 
Other methods tried to take advantage of the output produced by multiple trackers \cite{MEEM,Yoon2012,Wang2014,Bailer2014,HMMTxD}.
Thus, one can imagine that different trackers incorporate different knowledge, and this may constitute a valuable resource to leverage during tracking. 

Lately, the knowledge distillation (KD) framework \cite{Hinton2014KD} was introduced in the deep learning panorama as paradigm for, among the many \cite{He2016ResNet,Tang2016,Li2017KDnoise,Phuong2019}, knowledge transferring between models \cite{Geras2015} and model compression \cite{Chen2017,Howard2017,Polino2018}.
The idea boils down into considering a student model and one or more teacher models to learn from. Teachers explicit their knowledge through demonstrations on a never seen before transfer set. Through specific loss functions, the student is set to learn a task by matching the teachers' output and the ground-truth labels.
As visual tracking requires fast and accurate methods, KD can be a valuable tool to transfer the tracking ability of more accurate teacher trackers to more compact and faster student ones.
However, the standard setup of KD does not provide methods to exploit teachers online, but just offline. 
This makes this methodology unsuitable for tracking, which has been shown to benefit from both offline and online methods \cite{ATOM,DiMP,ECO,Yun2017,Chen2018}.
In contrast to such an issue, RL techniques offer established methodologies to optimize not only policies but also policy evaluation functions \cite{Watkins1992,Konda2000,Sutton2000,Mnih2013,Mnih2016}, which are then used to extract decision strategies. Along with this, RL also gives the possibility to maximize arbitrary and non-differentiable performance measures, and thus more tracking oriented objectives can be defined.

For the aforementioned motivations, the contribution of this paper is a novel tracking methodology where a student model exploits off-the-shelf trackers offline and online (tracking-by-trackers). The student is first trained via an effective strategy that combines KD and RL. After that, the model's compressed knowledge can be used interchangeably depending on the application's needs.
We will show how to exploit the student in three setups which result in, respectively, (i) a fast tracker (\myalgonamefirst), (ii) a tracker with a simple online mechanism (\myalgonamesecond), and (iii) a tracker capable of expert tracker fusion (\myalgonamethird). 
Through extensive evaluation procedures, it will be demonstrated that each of the algorithms competes with the respective state-of-the-art class of trackers while performing in real-time.

%% file: 2relatedwork.tex
\section{Related Work}

\subsubsection{Visual Tracking.} Here we review the trackers most related to ours. The network architecture implemented by the proposed student model takes inspiration from GOTURN \cite{GOTURN} and RE3 \cite{RE3}. These regression-based CNNs were shown to capture the target's motion while performing very fast. However, the learning strategy employed optimizes parameters just for coordinate difference. Moreover, great amount of data is needed to make such models achieve good accuracy. In contrast, our KD-RL-based method offers parameter optimization for overlap maximization and extracts previously acquired knowledge from other trackers requiring less labeled data.
Online adaptation methods like discriminative model learning \cite{TLD,MDNet,RealTimeMDNet} or discriminative correlation filters \cite{ECO,ATOM,DiMP} have been studied extensively to improve tracking accuracy. These procedures are time-consuming and require particular assumptions and careful design. We propose a simple online update strategy where an off-the-shelf tracker is used to correct the performance of the student model. Our method does not make any assumption on such tracker, and thus it can be freely selected to adapt to application needs.
Present fusion models exploit trackers in the form of discriminative trackers \cite{MEEM}, CNN feature layers \cite{Qi2016}, correlation filters \cite{Li2019} or out-of-the-box tracking algorithms \cite{Yoon2012,Wang2014,Bailer2014,HMMTxD}. However, such models work just online and do not take advantage of the great amount of offline knowledge that expert trackers can provide. Furthermore, they are not able to track objects without them. Our student model addresses these issues thanks to the decision making strategy learned via KD and RL.

\subsubsection{KD and RL.} We review the learning strategies most related to ours. KD techniques have been used for transferring knowledge between teacher and student models \cite{Bucila2006,Hinton2014KD}, where the supervised learning setting was employed more \cite{Geras2015,He2016ResNet,Tang2016,Li2017KDnoise} than the setup that uses RL \cite{Rusu2016,Parisotto2016}. 
In the context of computer vision, KD was employed for action recognition \cite{Garcia2018KDAR,Wank2019KDAR}, object detection \cite{Chen2017,Shmelkov2017}, semantic segmentation \cite{Liu2019semantic,He2019}, person re-identification \cite{Wu2019KDreid}. 
In the visual tracking panorama, KD was explored in \cite{Wang2019,Liu2019} to compress, CNN representations for correlation filter trackers and siamese network architectures, respectively. However, these works offer methods involving teachers specifically designed as correlation filter and siamese trackers, and so cannot be adapted to generic-approach visual trackers as we propose in this paper. Moreover, to the best of our knowledge, no method mixing KD and RL is currently present in the computer vision literature.
Our learning procedure is also related to the strategies that use deep RL to learn tracking policies \cite{Yun2017,Supancic2017,Ren2018,Chen2018,Meshgi2019}. Our formulation shares some characteristics with such methods in the markov decision process (MDP) definition, but our proposed learning algorithm is different as no present method leverages on teachers to learn the policy.

%% file: 3methods.tex
\section{Methodology}
The key point of this paper is to learn a simple and fast student model with versatile tracking abilities. KD is used for transferring the tracking knowledge of off-the-shelf trackers to a compressed model. 
However, as both offline and online strategies are necessary for tracking \cite{ATOM,DiMP,Yun2017,Chen2018}, we propose to augment the KD framework with an RL optimization objective. RL techniques deliver unified optimization strategies to directly maximize a desired performance measure (in our case the overlap between prediction and ground-truth bounding boxes) and to predict the expectation of such measure. We use the latter as base for an online evaluation and selection strategy. Put in other words, combining KD and RL lets the student model extract a tracking policy from teachers, improve it, and express its quality through an overlap-based objective.


\subsection{Preliminaries} 
Given a transfer set of videos $\dataset = \{ \video_0, \cdots,\video_{|\mathcal{D}|}\}$, we consider the $j$-th video
$\video_j = \big\{ \frame_t \in \images \big\}_{t=0}^{T_j}$
as a sequence of frames $\frame_t$, where $\images =  \{0,\cdots,255\}^{w \times h \times 3}$ is the space of RGB images. 
 Let $\bbox_t = [x_t,y_t,w_t,h_t] \in \reals^4$ be the $t$-th bounding box defining the coordinates of the top left corner, and the width and height of the rectangle that contains the target object.
At time $t$, given the current frame $\frame_t$, the goal of the tracker is to predict $\bbox_{t}$ that best fits the target in $\frame_{t}$.
We formally consider the student model as 
$\student : \images \rightarrow \reals^4 \times \reals$
that is a function which outputs the relative motion between $\bbox_{t-1}$ and $\bbox_t$, and the performance evaluation $\statevalue_t$, when inputted with frame $\frame_t$.
Similarly, we define the set of tracking teachers as
$\teachers = \big\{ \teacher : \images \rightarrow \reals^4 \big\}$
where each $\teacher$ is a function that, given a frame image, produces a bounding box estimate for that frame.

\begin{figure}[t]
\centering
\begin{minipage}[b]{.49\textwidth}
\centering
\includegraphics[width=.75\columnwidth]{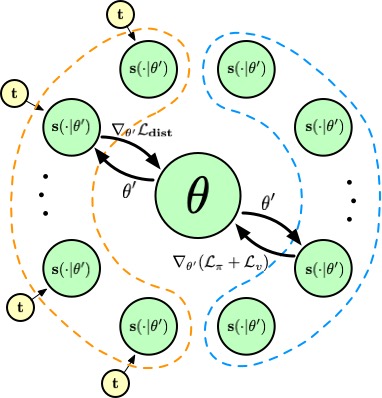}
\caption{Scheme of the proposed KD-RL-based learning framework. $S$ students interact independently with $\mdp_j$. After each $\episode_j$ done, a copy $\weights'$ of the shared weights $\weights$ is sent to each one. Every $t_{max}$ steps each student send the computed gradients to apply an update on $\weights$. The distilling students (highlighted by the orange dashed contour) extract knowledge from teachers $\teacher$ by optimizing $\distillationloss$. Autonomous students (circled with the blue dashed contour) learn an autonomous tracking policy by optimizing jointly $\policyloss$ and $\valueloss$.}
\label{fig:framework}
\end{minipage}%
\hfill
\begin{minipage}[b]{.49\textwidth}
\centering
\includegraphics[width=.9\columnwidth]{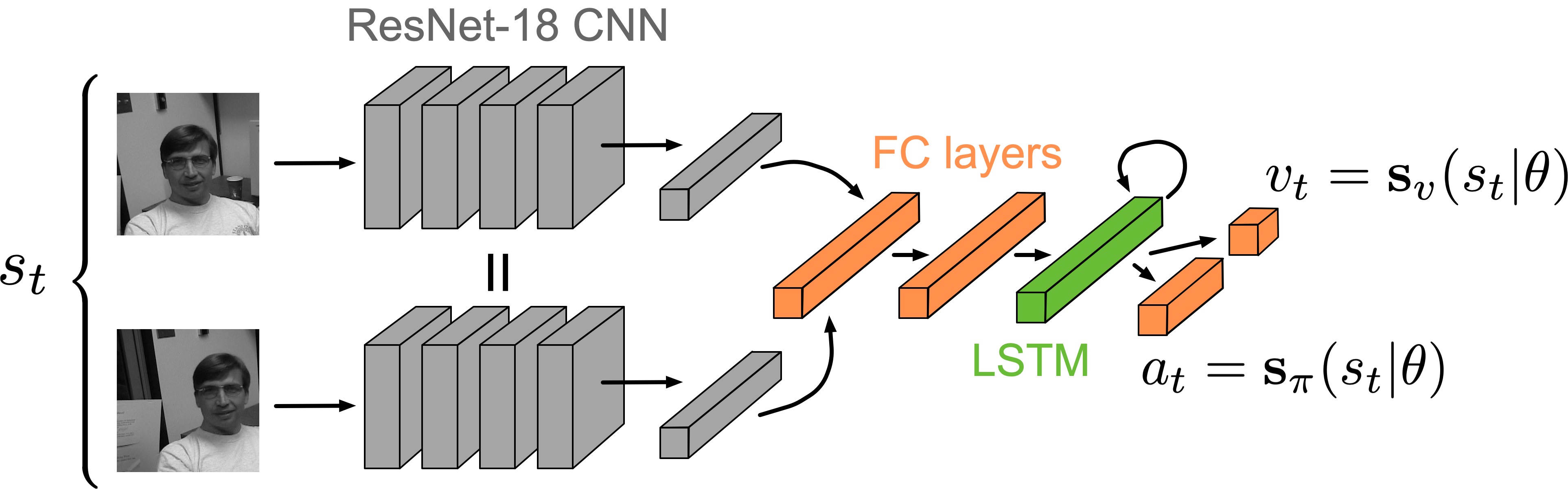}
   \caption{The student architecture is composed by two branches of convolutional layers (gray boxes) with shared weights followed by, two fully-connected layers (orange boxes), an LSTM layer (in green) and two parallel fully connected layers for the prediction of $v_t$ and  $a_t$ respectively.}
\label{fig:arch}
\includegraphics[width=.9\columnwidth]{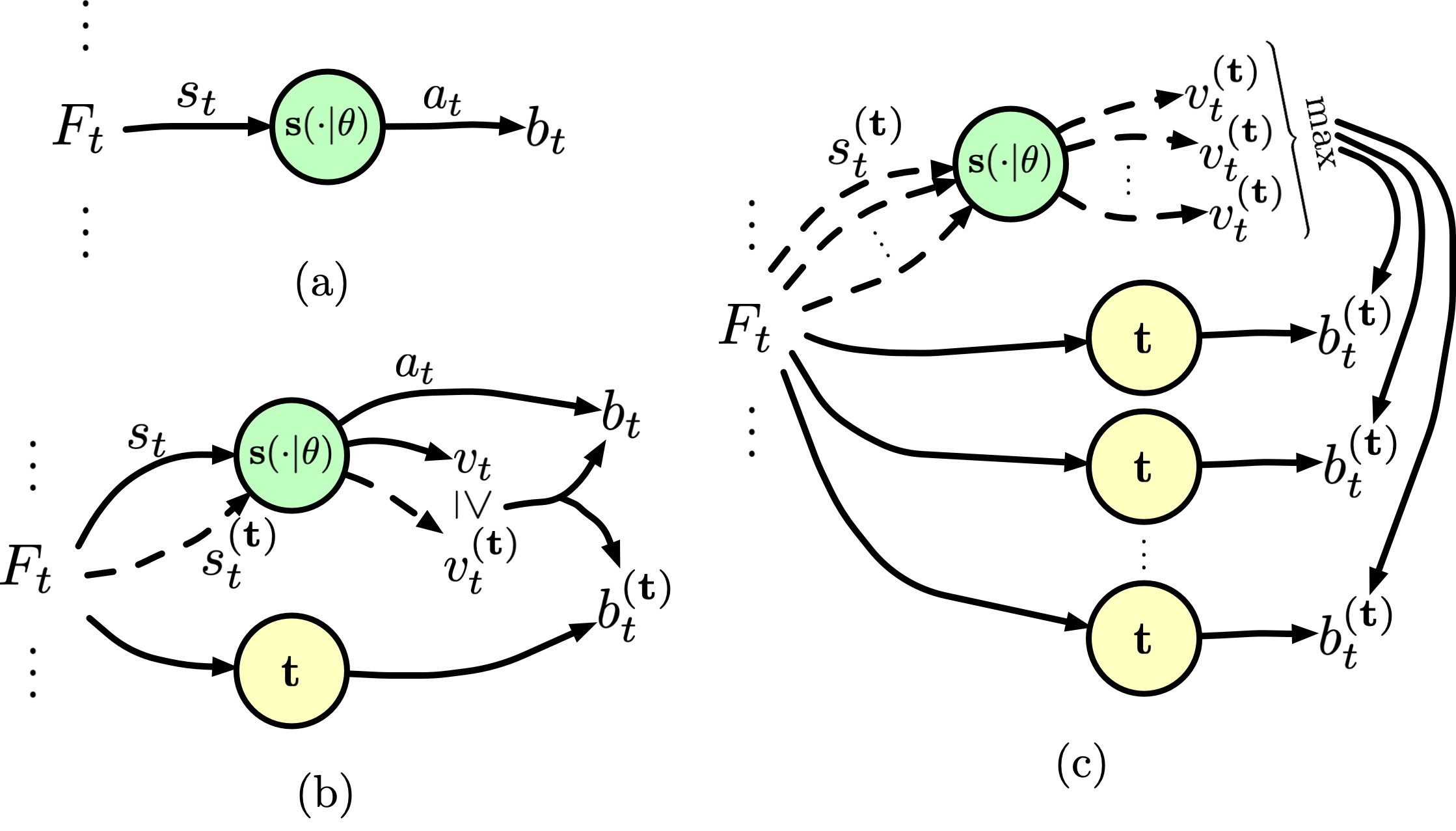}
\caption{Visual representation showing how the student and teachers are employed in the proposed trackers at every frame $\frame_t$. (a) represents \myalgonamefirst, (b) \myalgonamesecond, and (c) \myalgonamethird.}
\label{fig:trackers}
\end{minipage}
\end{figure}

\subsection{Visual Tracking as an MDP}
\label{sec:votmdp}
In our setting, $\student$ is treated as an artificial agent which interacts with an MDP defined over a video $\video_j$. 
The interaction happens through a temporal sequence of states $\state_1, \state_2, \cdots, \state_t \in \states$, actions $\action_1, \action_2, \cdots, \action_t \in \actions$ and rewards $\reward_1, \reward_2, \cdots, \reward_t \in [-1,1]$. In the $t$-th frame, the student is provided with the state $\state_t$ and outputs the continuous action $\action_t$ which consists in the relative motion of the target object, i.e. it indicates how its bounding box, which is known in frame $t-1$, should move to enclose the target in the frame $t$. $\action_t$ is rewarded by the measure of its quality $\reward_t$. We refer this interaction process as the episode $\episode_j$, which dynamics are defined by the MDP 
$\mdp_j = (\states, \actions, \reward, f)$.

\subsubsection{States.} Every $\state_t \in \states$ is defined as a pair of image patches obtained by cropping $\frame_{t-1}$ and $\frame_t$ using $\bbox_{t-1}$.
Specifically, $\state_t = \rho(\frame_{t-1}, \frame_t, \bbox_{t-1}, c)$, where $\rho(\cdot)$ crops the frames $\frame_{t-1}, \frame_t$ within the area of the bounding box $\bbox_{t-1}' = [x'_{t-1},y'_{t-1}, \contextfactor \cdot w_{t-1}, \contextfactor \cdot h_{t-1}]$ that has the same center coordinates of $\bbox_{t-1}$ but which width and height are scaled by $\contextfactor$. By selecting $\contextfactor > 1$, we can control the amount of additional image context information to be provided to the student. 

\subsubsection{Actions and State Transition.} Each $\action_t \in \actions$ consists in a vector $\action_t = [\Delta x_t, \Delta y_t, \allowbreak \Delta w_t, \Delta h_t] \in [-1,1]^4$ which defines the relative horizontal and vertical translations ($\Delta x_t, \Delta y_t$, respectively) and width and height scale variations ($\Delta w_t, \Delta h_t$, respectively) that have to be applied to $\bbox_{t-1}$ to predict $\bbox_t$. The latter step is obtained through $\psi: \actions \times \reals^4 \rightarrow \reals^4$.\footnote{Please refer to Appendix \ref{sec:mdpaux} for the definition of $\psi(\cdot)$.}
After performing $\action_t$, the student moves through $f$ from $\state_t$ to $\state_{t+1}$ which is defined as the pair of cropped images obtained from $\frame_t$ and $\frame_{t+1}$ using $\bbox_t$.

\subsubsection{Reward.} The reward function expresses the quality of $\action_t$ taken at $\state_t$ and it is used to feedback the student. 
Our reward definition is based on the Intersection-over-Union (IoU) metric computed between $\bbox_t$ and the ground-truth bounding box, denoted as $\bboxgt_t$, i.e., \begin{align}
\text{IoU}(\bbox_t, \bboxgt_t) = (\bbox_t \cap \bboxgt_t) / (\bbox_t \cup \bboxgt_t) \in [0,1].
\end{align}
At every interaction step $t$, the reward is formally defined as
\begin{align}
\reward_t = \reward(\bbox_t, \bboxgt_t) = 
    \begin{cases}
    \omega\left(\text{IoU}(\bbox_t, \bboxgt_t)\right) \ \text{if IoU}(\bbox_t, \bboxgt_t) \geq 0.5 \\
    -1 \ \text{otherwise}
    \end{cases}
\end{align}
with
$\omega(z) = 2(\lfloor z \rfloor_{0.05}) - 1$ that
floors to the closest $0.05$ digit and shifts the input range from $[0,1]$ to $[-1,1]$. 
 
\subsection{Learning Tracking from Teachers}
The student $\student$ is first trained in an offline stage. Through KD, knowledge is transferred from $\teachers$ to $\student$. By means of RL, such knowledge is improved and the ability of evaluating its quality is also acquired. All the gained knowledge will be used for online tracking.
We implement $\student$ as a parameterized function $\student(\state_t | \weights) : \states \rightarrow \actions \times \reals$ that given $\state_t$ outputs at the same time the action $a_t$ and state-value $v_t$. In RL terms, $\student$ maintains representations of both the policy $\student_{\pi} : \states \rightarrow \actions$ and the state value  $\student_{\statevalue} : \states \rightarrow \reals$ functions. 
The proposed learning framework, which is depicted in Figure \ref{fig:framework}, provides a single offline end-to-end learning stage. 
$S$ students are distributed as parallel and independent learning agents. Each one owns a set of learnable weights $\weights'$ that are used to generate experience by interacting with $\mdp_j$. The obtained experience,
in the form of $\nabla_{\weights'}$, is used to update asynchronously a shared set of weights $\weights$. After ending $\episode_j$, each student updates its $\weights'$ by copying the values of the currently available $\weights$. The entire procedure is repeated until convergence. This learning architecture follows the recent trends in RL that make use of distributed algorithms to speed up the training phase \cite{Gorila,Mnih2016,IMPALA}. 
We devote half of the students, which we refer to as distilling students, in acquiring knowledge from the teachers' tracking policy. The other half, called autonomous students, learn to track by interacting with $\mdp_j$ autonomously. 

\subsubsection{Distilling Students.} Each distilling student interacts with $M_j$ by observing states, performing actions and receiving rewards just as an autonomous student. However, to distill knowledge independently from the teachers' inner structure, we propose the student to learn from the actions of $\teacher \in \teachers$, which are executed in parallel. In particular, $\teacher$ is exploited every $t_{max}$ steps with the following loss function
\begin{align}
\label{eq:distloss}
\distillationloss &= \sum_{i=1}^{t_{max}} |a^{(\teacher)}_t - \student_{\pi}(\state_t | \weights')| \cdot m_i, 
\end{align}
which is the L1 loss between the actions performed by the student and the actions $a^{(\teacher)}_t = \phi(\bbox^{(\teacher)}_t, \bbox_{t-1})$ that the teacher would take to move the student's bounding box $\bbox_{t-1}$ into the teacher's prediction $\bbox^{(\teacher)}_t$.\footnote{Please refer to Appendix \ref{sec:mdpaux} for the definition of $\phi(\cdot)$.} At every $t$, $\teacher$ is selected as 
\begin{align}
\teacher \in \teachers \: : \: \IoU(\bbox^{(\teacher)}_t, \bboxgt_t) = \max_{\teacher \in \teachers}  \IoU(\bbox^{(\teacher)}_t, \bboxgt_t)
\end{align} 
as we would like to learn always from the best teacher.
The absolute values are multiplied by $m_i \in \{0,1\}$. Each of these is computed along the interaction and determines the status in which $\student$ performed worse than $\teacher$ ($m_i = 1$) or better ($m_i = 0$) in terms of the rewards $\reward(\state_t, \action_t)$ and $\reward(\state_t, \action^{(\teacher)}_t)$. The whole Eq. (\ref{eq:distloss}) is similar to what proposed in \cite{Chen2017} for KD from bounding-box predictions of object detectors. However, here we provide a temporal formulation of such objective and we swap the L2 loss with the L1, which was shown to work better for regression-based trackers \cite{GOTURN,RE3}.
By optimizing Eq. (\ref{eq:distloss}), the weights $\theta$ are changed only if the student's performance is lower than the performance of the teacher. In this way, we make the teacher transferring its knowledge by suggesting actions only in bad performing cases. In the others, we let the student free to follow its current tracking policy since it is superior. 

\subsubsection{Autonomous Students.} The learning process performed by the autonomous students follows the standard RL method for continuous control \cite{SuttonBarto2018}. Each student interacts with $M_j$ for a maximum of $t_{max}$ steps. At each step $t$, the students sample actions from a normal distribution $\mathcal{N}(\mu, \sigma)$, where the mean is defined as the student's predicted action, $\mu = \student_{\pi}(\state_t | \theta')$, and the standard deviation is obtained as $\sigma = |\student_{\pi}(\state_t | \theta') - \phi(\bboxgt_t, \bbox_{t-1}) |$ (which is the absolute value of the difference between the student's action and the action that obtains, by shifting $\bbox_{t-1}$, the ground-truth bounding box $\bboxgt_t$). Intuitively, $\mathcal{N}$ shrinks when $\action_t$ is close to the ground-truth action $\phi(\bboxgt_t, \bbox_{t-1})$, reducing the chance of choosing potential wrong actions when approaching the correct one. On the other hand, when $\action_t$ is distant from $\phi(\bboxgt_t, \bbox_{t-1})$, $\mathcal{N}$ spreads letting the student explore more. 
The students also predict $\statevalue_t = \student_{\statevalue}(\state_t| \weights)$ which is the cumulative reward that the student expects to receive from $\state_t$ to the end of the interaction. Since the proposed reward definition is a direct measure of the IoU occurring between the predicted and the ground-truth bounding boxes, $\student_{\statevalue}(\state_t | \weights)$ gives an estimate of the total amount of IoU that $\student$ expects to obtain from state $s_t$ on wards. Thus, this function can be exploited as a future-performance evaluator. 
After $t_{max}$ steps of interaction, the gradient to update the shared weights $\theta$ is built as
\begin{gather}
    \nabla_{\theta'} (\policyloss + \valueloss) \\
\label{eq:policyloss}
\policyloss = -\sum_{i=1}^{t_{max}} \log \student_{\pi}(\state_i | \theta')\big(r_i + \gamma \student_{\statevalue}(\state_{i+1} | \theta') - \student_{\statevalue}(\state_i | \theta') \big) \\
\label{eq:valueloss}
\valueloss = \sum_{i=1}^{t_{max}} \frac{1}{2} \big(R_i -  \student_{\statevalue}(\state_i | \theta') \big)^2, 
R_i = \sum_{k=1}^{i}\gamma^{k-1}\reward_k
\end{gather}
where (\ref{eq:policyloss}) is the policy loss and 
(\ref{eq:valueloss}) is the value loss. These definitions follow the standard advantage actor-critic objective \cite{Sutton2000}. 

To further facilitate and improve the learning, a curriculum learning strategy \cite{Bengio2009} is built for each parallel student. During learning, the length of the interaction is increased as $\student$ performs better than $\teachers$. Details are given in Appendix \ref{sec:curriculum}.

\subsection{Student Architecture}
The architecture used to maintain the representation of both the policy $\student_{\pi}$ and the state value $\student_{\statevalue}$ functions, which is pictured in Figure \ref{fig:arch}, is simple and presents a structure similar to the one proposed in \cite{GOTURN,RE3}. 
The network gets as input two image patches that pass through two ResNet-18 based \cite{He2016ResNet} convolutional branches that share weights. The feature maps produced by the branches are first linearized, then concatenated together and finally fed to two consecutive fully connected layers with ReLU activations. After that, features are given to an LSTM \cite{Hochreiter1997LSTM} layer. Both the fully connected layers and the LSTM are composed of 512 neurons. The output of the LSTM is ultimately fed to two separate fully connected heads, one that outputs the action $\action_t = \student_{\pi}(\state_t | \theta)$ and the other that outputs the value of the state $\statevalue_t = \student_{\statevalue}(\state_t | \theta)$.

\subsection{Tracking after Learning}
After the learning process, the student $\student(\cdot | \weights)$ is ready to be used for tracking. Here we describe three different ways in which $\student(\cdot | \weights)$ can be exploited:
\begin{enumerate}
    \item the student's learned policy $\student_{\pi}(\state_t | \weights)$ is used to predict bounding boxes $\bbox_t$ independently from the teachers. We call this setting \myalgonamefirst (TRAcking Student).
    \item the learned policy $\student_{\pi}(\state_t | \weights)$ and value function $\student_{\statevalue}(\state_t | \weights)$ are used to, respectively, predict $\bbox_t$ and evaluate $\student$ and $\teacher \in \teachers$ tracking behaviors, in order to correct the former's performance. We refer to this setup as \myalgonamesecond (TRAcking Student and Teacher).
    \item the learned state-value function $\student_{\statevalue}(\state_t | \weights)$ is used to evaluate the performance of the pool of teachers $\teachers$ in order to choose the best $b^{(\teacher)}_t$ and perform tracker fusion. We call this setup \myalgonamethird (TRAcking by Student FUSing Teachers).
\end{enumerate}

In the following, we provide more details about the three settings. For a better understanding, the setups are visualized in Figure \ref{fig:trackers}.

\subsubsection{\myalgonamefirst.} In this setting, each tracking sequence $\mathcal{V}_j$, with target object outlined by $\bboxgt_0$, is considered as $\mdp_j$ described in section \ref{sec:votmdp}. States $\state_t$ are extracted from frames $\frame_{t-1}, \frame_t$, actions are performed by means of the student's learned policy $\action_t = \student_{\pi}(\state_t | \weights)$ and are used to output the bounding boxes $\bbox_t = \psi(\action_t, \bbox_{t-1})$. 
This setup is fast as it requires just a forward pass through the network to obtain a bounding box prediction.

\subsubsection{\myalgonamesecond.} 
In this setup, the student makes use of the learned $\student_{\pi}(\state_t | \weights)$ to predict $\bbox_t$ and $\student_{\statevalue}(\state_t | \weights)$ to evaluate its running tracking quality and the one of $\teacher \in \teachers$ which is run in parallel. 
In particular, at each time step $t$, $\statevalue_t = \student_{\statevalue}(\state_t | \weights)$ and $\statevalue^{(\teacher)}_t = \student_{\statevalue}(\state^{(\teacher)}_t | \weights)$ are obtained as performance evaluation for $\student$ and $\teacher$ respectively. The teacher state is obtained as $\state^{(\teacher)}_t = \rho(\frame_{t-1}, \frame_t, \bbox^{(\teacher)}_{t-1}, \contextfactor)$. By comparing the two expected returns, \myalgonamesecond decides if to output the student's or the teacher's bounding box. More formally, if $\statevalue_t \geq \statevalue^{(\teacher)}_t$ then $b_t := \psi(\action_t, \bbox_{t-1})$ otherwise $\bbox_t := \bbox^{(\teacher)}_{t}$. This assignment has the side effect of correcting the tracking behaviour of the student as, at the successive time step, the previously known bounding box becomes the previous prediction of the teacher. 
Thus, the online adaption consists in a very simple procedure that evaluates $\teacher$'s performance to eventually pass control to it. Notice that, at every $t$, the execution of $\teacher$ is independent from $\student$ as the second does not need the first to finish because the evaluations are done based on the predictions given at $t-1$. Hence, the executions of the two can be put in parallel, with the overall speed of \myalgonamesecond resulting is the lowest between the one of $\student$ and $\teacher$. 

\subsubsection{\myalgonamethird.} 
\label{sec:trasfust}
In this tracking setup, just the student's learned state-value function $\student_{\statevalue}(\state_t | \weights)$ is exploited. At each step $t$, teachers $\teacher \in \teachers$ are executed following their standard methodology. States $\state^{(\teacher)}_t = \rho(\frame_{t-1}, \frame_t, \bbox^{(\teacher)}_{t-1}, \contextfactor) \: \forall \teacher \in \teachers$ are obtained. The performance evaluation of the teachers is obtained through the student as $\statevalue^{(\teacher)}_t = \student_{\statevalue}(\state^{(\teacher)}_t | \theta)$. The output bounding box is selected as $\bbox_t = \bbox^{(\teacher')}$ by considering the teacher $\teacher'$ that achieves the highest expected return, i.e. 
\begin{align}
\teacher' \in \teachers \: : \: \statevalue^{(\teacher')}_t = \max_{\teacher \in \teachers} \statevalue^{(\teacher)}_t.
\end{align} 
This procedure consists in fusing sequence-wise the predictions of $\teachers$ and, similarly as for \myalgonamesecond, the execution of teachers and student can be put in parallel. In such setting, the speed of \myalgonamethird results in the lowest between the ones of $\student$ and of each $\teacher \in \teachers$.

%% file: 4experimental.tex
\input{tables/demostats.tex}

\section{Experimental Results}

\subsection{Experimental Setup}

\subsubsection{Teachers.}
The tracking teachers selected for this work are KCF \cite{KCF}, MDNet \cite{MDNet}, ECO \cite{ECO}, SiamRPN \cite{SiamRPN}, ATOM \cite{ATOM}, and DiMP \cite{DiMP}, due to their established popularity in the visual tracking panorama. Moreover, since they tackle visual tracking by different approaches, they can provide knowledge of various quality. 
In experiments, we considered exploiting single teacher or a pool of teachers. In particular, the following sets of teachers were examined $\tk = \{ \tkcf\}, \tm = \{ \tmdnet \},  \te = \{ \teco \}, \ts = \{ \tsiamrpn \}, \ta = \{ \tatom \}, \td = \{ \tdimp \}, \tp = \{ \tkcf, \tmdnet, \teco, \tsiamrpn \}$.

\subsubsection{Transfer Set.}
The selected transfer set was the training set of GOT-10k dataset \cite{GOT10k}, due to its large scale. 
Just $\tk, \tm, \te, \ts$ were used for offline learning, as none of these was trained on this dataset. This is an important point because unbiased examples of the trackers' behavior should be exploited to train the student.
Moreover, predictions that exhibit meaningful knowledge should be retained. Therefore, we filtered out all the videos $\video_j$ which teacher predictions did not satisfy  $\text{IoU}(\bbox^{(\teacher)}_t, \bboxgt_t) > \dfiltcoeff$ for all $t \in \{1,\dots,T_j\}$. We considered $\dfiltcoeff = 0.5$ as minimum threshold for a prediction to be considered positive, and we then varied $\dfiltcoeff$ among 0.6, 0.7, 0.8, 0.9 for more precise predictions.
To produce more training samples, videos, and filtered trajectories were split in five randomly indexed sequences of 32 frames and bounding boxes, similarly as done in \cite{RE3}.
In Table \ref{tab:demostats} a summary of $\dataset$ is presented. The number of positive trajectories, the average overlap (A0) \cite{GOT10k} on the transfer set, and the total number of sequences $|\dataset|$ are reported per teacher and per $\dfiltcoeff$.

\subsubsection{Benchmarks and Performance Measures.}
We performed performance evaluations on the GOT-10k test set \cite{GOT10k}, UAV123 \cite{UAV123}, LaSOT \cite{LaSOT}, OTB-100 \cite{OTB} and VOT2019 \cite{VOT2019} datasets. These offer videos of various nature and difficulty, and are all popular benchmarks in the visual tracking community.
The evaluation protocol used for GOT-10k is the one-pass evaluation (OPE) \cite{OTB}, along with the metrics: AO, and success rates (SR) with overlap thresholds $0.50$ and $0.75$. For UAV123, LaSOT, and OTB-100 the OPE method was considered with the area-under-the-curve (AUC) of the success and precision plot, referred to as success score (SS) and precision scores (PS) respectively \cite{OTB}. Evaluation on VOT2019 is performed in terms of expected average overlap (EAO), accuracy (A), and robustness (R) \cite{VOT}.
Further details about the benchmarks are given in Appendix \ref{sec:benchmarks}.

\subsubsection{Implementation Details.}
The image crops of $\state_s$ were resized to $[128 \times 128 \times 3]$ pixels and standardized by the mean and standard deviation calculated on the ImageNet dataset \cite{ImageNet}. The ResNet-18 weights were pre-trained for image classification on the same dataset \cite{ImageNet}. The image context factor $\contextfactor$ was set to $1.5$. The training videos were processed in chunks of 32 frames. At test time, every 32 frames, the LSTM's hidden state is reset to the one obtained after the first student prediction (i.e. $t = 1$), following \cite{RE3}.
Due to hardware constraints, a maximum of $S = 24$ training students were distributed on 4 NVIDIA TITAN V GPUs of a machine with an Intel Xeon E5-2690 v4 @ 2.60GHz CPU and 320 GB of RAM. 
The discount factor $\gamma$ was set to 1. The length of the interaction before an update was defined in $t_{max}= 5$ steps. 
The Radam optimizer~\cite{Radam} was employed and the learning rate for both distilling and autonomous students was set to $10^{-6}$. A weight decay of $10^{-4}$ was also added to $\distillationloss$ as regularization term. To control the magnitude of the gradients and stabilize learning, $(\policyloss + \valueloss)$ was multiplied by $10^{-3}$.
The student was trained until the validation performance on the GOT-10k validation set stopped improving. Longest trainings took around 10 days.
The speed of the parallel setups of \myalgonamesecond and \myalgonamethird was computed by considering the speed of the slowest tracker (student or teacher) plus an overhead.
Code was implemented in Python and is available here\footnote{\url{https://github.com/dontfollowmeimcrazy/vot-kd-rl}}.
Source code publicly available was used to implement the teacher trackers. Default configurations were respected as much as possible. For a fair comparison, we report the results of such implementations, that have slightly different performance than stated in the original papers.

%% file: tables/demostats.tex
    
    
	
	
	
	
	


\begin{table*}[t]
\fontsize{5}{6}\selectfont
	\centering
	\caption{Teacher-based statistics of the transfer set.}
	\label{tab:demostats}
	\setlength\tabcolsep{.05cm}
	\begin{tabular}{l | c c c | c c c | c c c | c c c | c c c }
		\toprule

		Teachers & \multicolumn{3}{c|}{$\dfiltcoeff = 0.5$} & \multicolumn{3}{c|}{$\dfiltcoeff = 0.6$} & \multicolumn{3}{c|}{$\dfiltcoeff = 0.7$} &  \multicolumn{3}{c|}{$\dfiltcoeff = 0.8$}  &  \multicolumn{3}{c}{$\dfiltcoeff= 0.9$}  \\

		 & \# traj & AO & $|\mathcal{D}|$ & \# traj & AO & $|\mathcal{D}|$  & \# traj & AO & $|\mathcal{D}|$ & \# traj & AO & $|\mathcal{D}|$ & \# traj & AO & $|\mathcal{D}|$ \\
		\midrule
		$\tk$ & 1884 &	0.798 & 9225 & 1097 & 0.836 & 5349 & 439 & 0.873 & 2122 & 73 & 0.914 & 356 & 0 & 0.0 & 0  \\
		$\tm$ & 1600 & 0.767 & 7859 & 781 & 0.808 & 3831 & 216 & 0.851 & 1052 & 18 & 0.898 & 86 & 0 & 0.0 & 0  \\
		$\te$ & 2754 & 0.808 & 13526 & 1659	& 0.843	& 8122	& 720 & 0.879 & 3507 & 160 & 0.915 & 773 & 1 & 0.954 & 4 \\
		$\ts$ & 3913 & 0.829 & 19259 & 2646	& 0.854	& 12997	& 1447 & 0.878 & 7080 & 431 & 0.908 & 2097 & 9 & 0.947 & 42  \\
		$\tp$ & 4519	& 0.840	& 22252	& 3092 & 0.863 & 15195 & 1698 & 0.887 & 8307 & 496 & 0.915 & 2414 & 10 & 0.948 & 46 \\
		\bottomrule		
\end{tabular}
\end{table*}

%% file: 5results.tex
\input{tables/ablation.tex}

\begin{figure}[t]
\begin{minipage}[t]{.49\textwidth}
\centering
\includegraphics[width=\columnwidth]{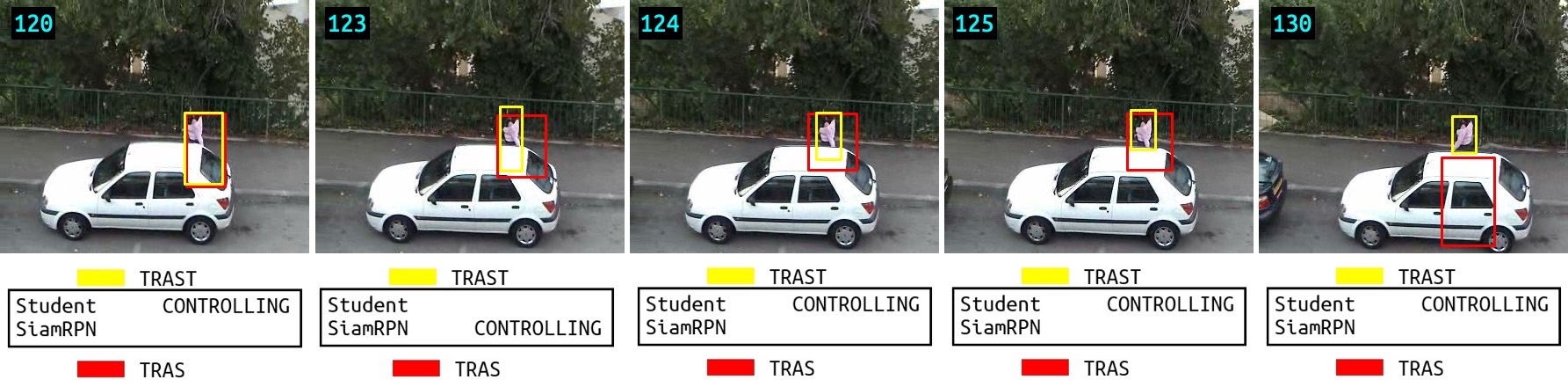}
\caption{Visual example of how \myalgonamesecond relies effectively on the teacher, passing control to $\ts$ and saving the simple student (\myalgonamefirst) from the drift.}
\label{fig:trastdecision}
\end{minipage}%
\hfill
\begin{minipage}[t]{.49\textwidth}
\centering
\includegraphics[width=.9\columnwidth]{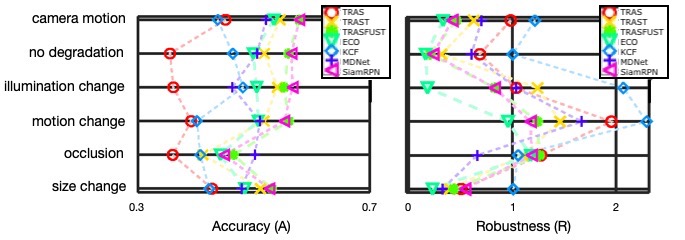}
\caption{Analysis of the accuracy (A) and robustness (R) on VOT2019 over different classes of tracking sequences. 
}
\label{fig:vot2019ar}
\end{minipage}
\end{figure}

\subsection{Results}
In the following sections, when not specified, the three tracker setups regard the student trained using $\tp$ and $\dfiltcoeff = 0.5$, paired with $\ts$ in \myalgonamesecond, and managing $\tp$ in \myalgonamethird.

\subsubsection{General Remarks.}
In Table \ref{tab:ablation} the performance of \myalgonamefirst, \myalgonamesecond, \myalgonamethird are reported, while the performances of the teachers are presented in the first six rows of Table \ref{tab:sota}. \myalgonamefirst results in a very fast method with good accuracy. Combining KD and RL results in the best performance, outperforming the baselines that use for training just the ground-truth (\myalgonamefirst-GT), KD and ground-truth (\myalgonamefirst-KD-GT), and just KD (\myalgonamefirst-KD). We did not report the performance of $\student$ trained only by RL because convergence was not attained due to the large state and action spaces. 
Benefiting the teacher during tracking is an effective online procedure. Indeed, \myalgonamesecond improves \myalgonamefirst by 24\% on average, and a qualitative example of the ability to pass control to the teacher is given in Figure \ref{fig:trastdecision}.
The performance of \myalgonamethird confirms the student's evaluation ability. This is the most accurate and robust tracker thanks to the effective fusion of the underlined trackers. 
Overall, all three trackers show balanced performance across the benchmarks, thus demonstrating good generalization.
No use of the curriculum learning strategy (\myalgonamefirst-no-curr,  \myalgonamesecond-no-curr, \myalgonamethird-no-curr) slightly decreases the performance of all. 
In Figure \ref{fig:vot2019ar} the performance of the trackers is reported for different classes of sequences of VOT2019, while in Figure \ref{fig:qualitativeex} some qualitative examples are presented. \footnote{For more, please see \url{https://youtu.be/uKtQgPk3nCU}}
These results demonstrate the effectiveness of our methodology and that the proposed student model respects, respectively, the goals (i), (ii), (iii) introduced in Section \ref{sec:intro}.

\input{tables/perteacher.tex}

\subsubsection{Impact Of Teachers.}
\begin{figure}[t]
\begin{minipage}[t]{.49\textwidth}
\centering
\includegraphics[width=\columnwidth]{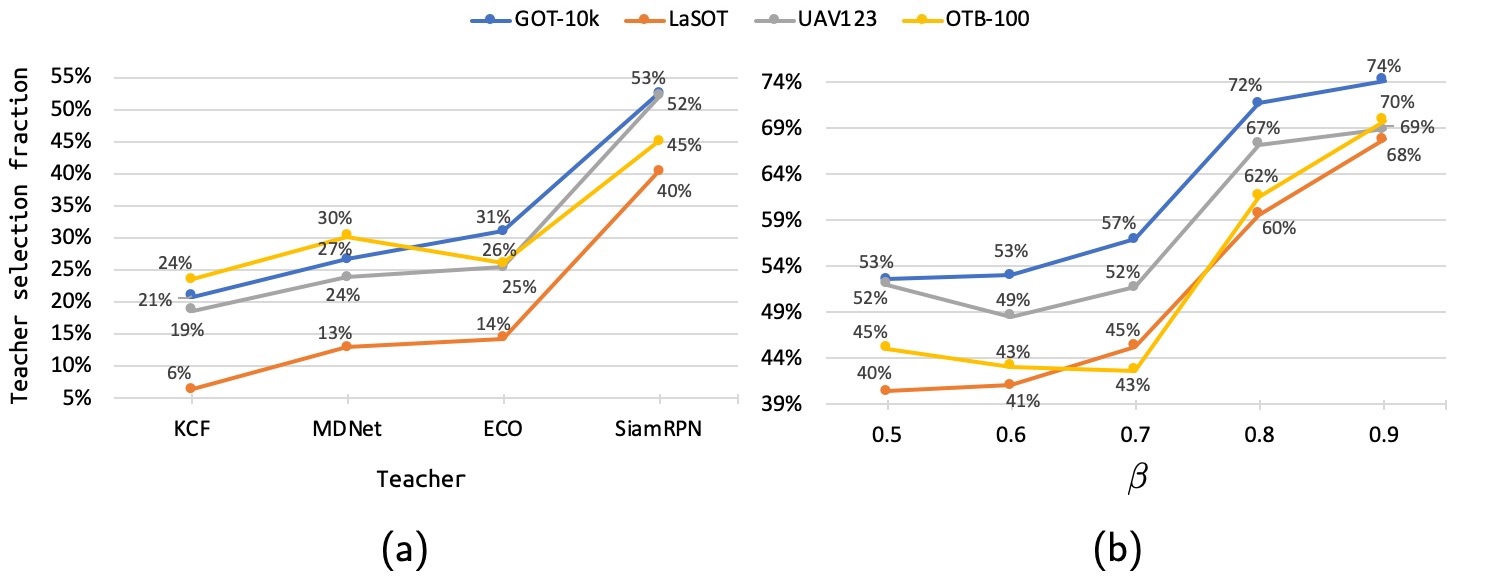}
\caption{Per benchmark fractions of predictions attributed to $\teacher$ in the \myalgonamesecond setup.}
\label{fig:decisionsline}
\includegraphics[width=\columnwidth]{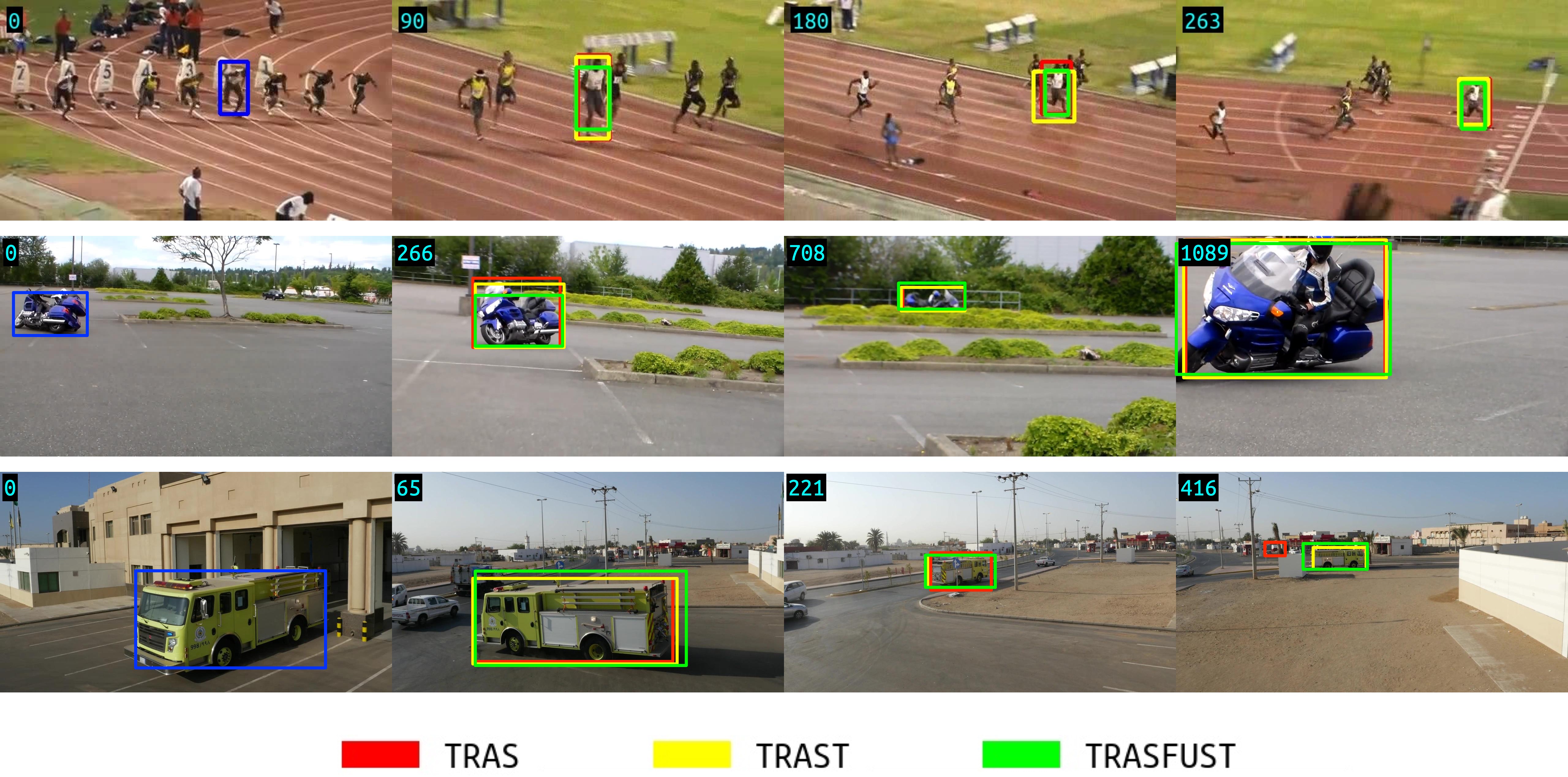}
\caption{Qualitative examples of the proposed trackers.}
\label{fig:qualitativeex}
\end{minipage}%
\hfill
\begin{minipage}[t]{.49\textwidth}
\centering
\includegraphics[width=.7\columnwidth]{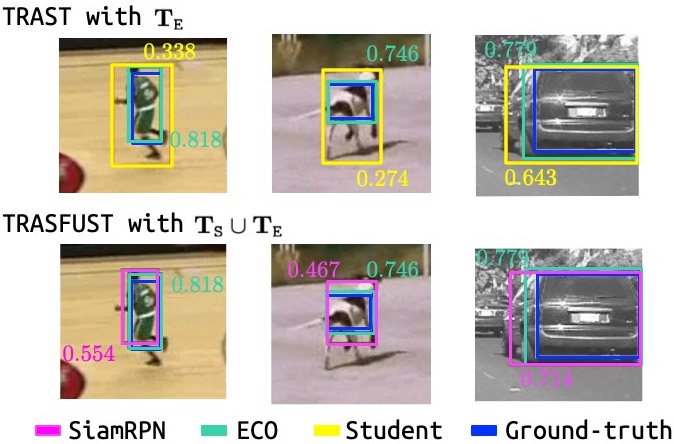}
\caption{Behaviour of \myalgonamesecond and \myalgonamethird with ambiguous ground-truths. In the presented frames, \myalgonamesecond selects the bounding box predicted by the student, while \myalgonamethird to one given by $\ts$. Those outputs are qualitative better but have much less $\IoU$ (quantified by the colored numbers) with respect to $\bboxgt_t$. This impacts the overall quantitative performance. }
\label{fig:qualerror}
\end{minipage}
\end{figure}

In Table \ref{tab:perteacher} the performance of the proposed trackers in different student-teacher setups is reported.
The general trend of the three trackers reflects the increasing tracking capabilities of the teachers. Indeed, on every considered benchmark, the tracking ability of the student increases as a stronger teacher is employed.
For \myalgonamesecond, this is also proven by Figure \ref{fig:decisionsline} (a), where we show that better teachers are exploited more.
Moreover, using more than one teacher during training leads to superior tracking policies and to  better exploitation of them during tracking.
Although in general student models cannot outperform their teachers due to their simple and compressed architecture \cite{Cho2019}, \myalgonamefirst and \myalgonamesecond show such behavior on benchmarks where teachers are weak. 
Using two teachers during tracking is the best \myalgonamethird configuration, as in this setup it outperforms the best teacher by more than 2\% on all the considered benchmarks. 
When weaker teachers are added to the pool, the performance tends to decrease, suggesting a behavior similar to the one pointed out in \cite{Bailer2014}.
Part of the error committed by \myalgonamesecond and \myalgonamethird on benchmarks like OTB-100 and VOT2019 is explained by Figure \ref{fig:qualerror}. In situations of ambiguous ground-truth, such trackers make predictions that are qualitatively better but quantitatively worse.
\myalgonamesecond and \myalgonamethird show to be unbiased to the training teachers, as their capabilities generalize also to $\ta$ and $\td$ which are not exploited during training.
\input{tables/demofilt.tex}
In Table \ref{tab:demofilt} we present the performance of the proposed trackers while considering different quality of teacher actions. Increasing the quality, thus reducing the number of videos, results in decreasing the performance of all three trackers. The loss is not significant between $\dfiltcoeff = 0.5$ and $\dfiltcoeff = 0.7$, while considering more precise actions, \myalgonamefirst suffers majorly, suggesting that more data is a key factor for an autonomous tracking policy. Interestingly, \myalgonamesecond and \myalgonamethird are able to perform tracking even if the student is trained with limited training samples. The plot (b) in Figure \ref{fig:decisionsline} confirms that the student relies effectively to its teacher, as the latter's output is selected more often as $\student$ loses performance.

Running the student takes just 11ms on our machine. \myalgonamefirst performs at 90 FPS. The speed of \myalgonamesecond and \myalgonamethird depends on the chosen teacher and varies between 5 and 40 FPS, as shown in Table \ref{tab:perteacher}. In parallel setups, \myalgonamesecond and \myalgonamethird run in real-time if the teachers do so. 

\subsubsection{State of the Art Comparison.}
In Table \ref{tab:sota} we report the results of the proposed trackers against the state-of-the-art. In the following comparisons, we consider the results of the best configurations proposed in the above analysis.

\myalgonamefirst outperforms GOTURN and RE3 which employ a similar DNN architecture but different learning strategies. On GOT-10k and LaSOT it also surpasses the recent GradNet and ROAM, and GCT on UAV123.
\myalgonamesecond outperforms ATOM and SiamCAR on GOT-10k, UAV123, LaSOT, while losing little performance to DiMP. The performance is better than RL-based trackers \cite{Yun2017,Chen2018,Ren2018} on UAV123 and comparable on OTB-100. 
Finally, \myalgonamethird outperforms all the trackers on all the benchmarks (where the pool $\te \smallcup \ts$ was used). Remarkable results are obtained on UAV123 and OTB-100, with SS of 0.679 and 0.701 and PS of 0.873 and 0.931, respectively. Large improvement is achieved over all the methodologies that include expert trackers in their methodology.

\input{tables/sota2.tex}

%% file: tables/ablation.tex
\begin{table*}[t]
\fontsize{5}{6}\selectfont
	\centering
	\caption{Performance of the proposed trackers. Results of removing some components of our methodology are also reported. Best values, per contribution, are highlighted in red.}
	\label{tab:ablation}
	\setlength\tabcolsep{.08cm}
	\begin{tabular}{ l | c c c | c c | c c | c c }
		\toprule
		& \multicolumn{3}{c|}{GOT-10k} & \multicolumn{2}{c|}{UAV123} &  \multicolumn{2}{c|}{LaSOT} & \multicolumn{2}{c}{OTB-100}  \\

		\multirow{-2}{*}{Contribution} & AO & SR$_{0.50}$ & SR$_{0.75}$ & SS & PS & SS & PS & SS & PS  \\
		\midrule
		\myalgonamefirst-GT & 0.444 & 0.495 & 0.286 & 0.483 & 0.616 & 0.331 & 0.271 & 0.438 & 0.581 \\
		\myalgonamefirst-KD-GT & 0.448 & 0.499 & 0.305 & 0.491 & 0.630 & 0.354 & 0.298 & 0.448 & 0.606 \\
		\myalgonamefirst-KD & 0.422 & 0.481 & 0.239 & 0.494 & 0.634 & 0.340 & 0.276 & 0.457 & 0.635 \\
		\myalgonamefirst-no-curr & 0.474 & 0.547 & 0.307 & 0.501 & 0.644 & 0.385 & 0.323 & 0.447 & 0.600 \\
		\myalgonamefirst & 0.484 & 0.556 & 0.326 & 0.515 & 0.655 & 0.386 & 0.330 & 0.481 & 0.644 \\
		\myalgonamesecond-no-curr & 0.530 & \tblbest{0.630} & \tblbest{0.347} & 0.602 & 0.770 & 0.484 & 0.464 & 0.595 & 0.794 \\
		\myalgonamesecond & \tblbest{0.531} & 0.626 & 0.345 & 0.603 & 0.773 & 0.490 & 0.470 & 0.604 & 0.818 \\
		\myalgonamethird-no-curr & 0.506 & 0.599 & 0.278 & 0.627 & 0.819 & 0.496 & 0.484 & \tblbest{0.665} & 0.879 \\
		\myalgonamethird & 0.519 & 0.616 & 0.287 & \tblbest{0.628} & \tblbest{0.823} & \tblbest{0.510} & \tblbest{0.505} & 0.660 & \tblbest{0.890} \\
		\bottomrule		
\end{tabular}
\end{table*}

%% file: tables/perteacher.tex
\begin{table*}[t]
	\fontsize{5}{5.5}\selectfont
	\centering
	\caption{Performance of the proposed trackers while considering different teacher setups for training and tracking. Best results per tracker are highlighted in red, second-best in blue.}
	\label{tab:perteacher}
	\setlength\tabcolsep{.08cm}
	\begin{tabular}{ c | l | l | c c c | c c | c c | c c | c}
		\toprule
					& Training & Tracking & \multicolumn{3}{c|}{GOT-10k} & \multicolumn{2}{c|}{UAV123} & \multicolumn{2}{c|}{LaSOT} &  \multicolumn{2}{c|}{OTB-100} & \multirow{2}{*}{FPS} \\
		& Teachers & Teachers & AO & SR$_{0.50}$ & SR$_{0.75}$ & SS & PS  & SS  & PS & SS & PS  &  \\
		\midrule
		
		\scalebox{.75}{\parbox[t]{2mm}{\multirow{4}{*}{\rotatebox[origin=c]{90}{\myalgonamefirst}}}} & $\tk$ & - & 0.371 & 0.418 & 0.178 & 0.464 & 0.598 & 0.321 & 0.241 & 0.390 & 0.524 & \multirow{4}{*}{90} \\
		& $\tm$ 	& - & 0.414 & 0.473 & 0.214 & 0.462 & 0.606 & 0.336 & 0.262 & 0.390 & 0.545 & \\
		& $\te$	& - & 0.422 & 0.484 & 0.232 & 0.507 & \tblsecondbest{0.652} & 0.357 & 0.286 & 0.422 & 0.567 & \\
		& $\ts$	& - & \tblsecondbest{0.441} & \tblsecondbest{0.499} & \tblsecondbest{0.290} & \tblbest{0.517} & 0.646 & \tblsecondbest{0.377} & \tblsecondbest{0.310} & \tblsecondbest{0.447} & \tblsecondbest{0.599} & \\
		& $\tp$		& - & \tblbest{0.484} & \tblbest{0.556} & \tblbest{0.326} & \tblsecondbest{0.515} & \tblbest{0.655} & \tblbest{0.386} & \tblbest{0.330} & \tblbest{0.481} & \tblbest{0.644} & \\
		
	    \midrule
		
		\scalebox{.75}{\parbox[t]{2mm}{\multirow{9}{*}{\rotatebox[origin=c]{90}{\myalgonamesecond}}}} & $\tk$  & $\tk$ 		& 0.390 & 0.440 & 0.191 & 0.526 & 0.682 & 0.388 & 0.319 & 0.495 & 0.660 & 90 \\
		& $\tm$  & $\tm$   & 0.452 & 0.521	& 0.223 & 0.572	& 0.776 & 0.433 & 0.386 & 0.569	& 0.793 & 5 \\
		& $\te$ & $\te$  & 0.491	& 0.571	& 0.249	& 0.580	& 0.768	& 0.442	& 0.397	& 0.583	& 0.786 & 15 \\
		& $\ts$ 	& $\ts$  & 0.532 &	0.632 & 0.354 & 0.605 & 0.779 & 0.485 & 0.457 & 0.601 & 0.806 & 40 \\
		& $\tp$  & $\tk$		& 0.469 & 0.541	& 0.297	& 0.562	& 0.727	& 0.422 & 0.376 & 0.560	& 0.760 & 90 \\
		& $\tp$ & $\tm$		& 0.494	& 0.573	& 0.302	& 0.604	& 0.798	& 0.466 & 0.431 & 0.596 &	0.815 & 5 \\
		& $\tp$ & $\te$		& 0.521	& 0.607	& 0.307	& 0.606	& 0.795	& 0.456 & 0.419 & 0.608 & 0.822 & 15 \\
		& $\tp$ & $\ts$	&   0.531	& 0.626	& 0.345	& 0.603	& 0.773	& 0.490 & 0.470 & 0.604 & 0.818 & 40 \\
		& $\tp$  & $\ta$		& \tblsecondbest{0.557} & \tblsecondbest{0.640}	& \tblsecondbest{0.393}	& \tblsecondbest{0.634}	& \tblsecondbest{0.823}	& \tblsecondbest{0.513} & \tblsecondbest{0.488} & \tblsecondbest{0.623}	& \tblsecondbest{0.838} & 20 \\
		& $\tp$  & $\td$		& \tblbest{0.604} & \tblbest{0.708}	& \tblbest{0.469}	& \tblbest{0.647}	& \tblbest{0.837}	& \tblbest{0.545} & \tblbest{0.524} & \tblbest{0.643}	& \tblbest{0.865} & 25 \\
		
		\midrule
		
		\scalebox{.75}{\parbox[t]{2mm}{\multirow{1}{*}{\rotatebox[origin=c]{90}{\myalgonamethird}}}} & $\tp$ & $\tk \smallcup \tm$ & 0.317	& 0.319 & 0.105	& 0.493 & 0.720	& 0.396 & 0.372	& 0.666 & \tblsecondbest{0.901} & 5 \\
		& $\tp$ & $\tm \smallcup \te$ & 0.384	& 0.398 & 0.131	& 0.563 & 0.791	& 0.422 & 0.392	& \tblbest{0.701} & \tblbest{0.931} & 5 \\
		& $\tp$ & $\te \smallcup \ts$ & \tblsecondbest{0.526}	& \tblsecondbest{0.624} & \tblsecondbest{0.305}	& \tblsecondbest{0.634} & 0.815	& 0.507 & 0.500	& 0.670 & 0.877 & 15 \\
		& $\tp$ & $\ta \smallcup \td$ & \tblbest{0.617} & \tblbest{0.729}	& \tblbest{0.490}	& \tblbest{0.679} & \tblbest{0.873}	& \tblbest{0.576}	& \tblbest{0.574}	& \tblsecondbest{0.692} &	0.895 & 20 \\
		& $\tp$ & $\tm \smallcup \te \smallcup \ts$ & 0.517	& 0.615 & 0.294	& 0.633 & \tblsecondbest{0.823}	& \tblsecondbest{0.513} & 0.504	& 0.682 & 0.897 & 5 \\
		 & $\tp$ & $\tp$ & 0.519 & 0.616	& 0.287	& 0.628 & \tblsecondbest{0.823}	& 0.510	& \tblsecondbest{0.505}	& 0.660 & 0.890 & 5 \\
		\bottomrule		
\end{tabular}
\end{table*}

%% file: tables/demofilt.tex
\begin{table*}[t]
\fontsize{5}{5.5}\selectfont
	\centering
	\caption{Results of the proposed trackers considering $\tp$'s increasingly better predictions. Best values, per tracker, are highlighted in red, second-best in blue.}
	\label{tab:demofilt}
	\setlength\tabcolsep{.08cm}
	\begin{tabular}{l | l | c c c | c c | c c | c c }
		\toprule
		& & \multicolumn{3}{c|}{GOT-10k} & \multicolumn{2}{c|}{UAV123} & \multicolumn{2}{c|}{LaSOT} &  \multicolumn{2}{c}{OTB-100}  \\
		& \multirow{-2}{*}{Tracker} 	& AO & SR$_{0.50}$ & SR$_{0.75}$ & SS & PS  & SS  & PS & SS & PS \\
		\midrule
		
		 & \myalgonamefirst	& \tblbest{0.484} & \tblbest{0.556} & \tblbest{0.326} & \tblbest{0.515} & \tblbest{0.655} & \tblbest{0.386} & \tblbest{0.330} & \tblbest{0.481} & \tblbest{0.644} \\
		 & \myalgonamesecond	& \tblbest{0.532} & \tblbest{0.632} & \tblbest{0.354}	&  \tblbest{0.605} & \tblbest{0.779} & \tblbest{0.485} & \tblbest{0.457} & \tblsecondbest{0.601} & \tblsecondbest{0.806} \\
		 \multirow{-3}{*}{$\dfiltcoeff = 0.5$} & \myalgonamethird	& \tblbest{0.519}	& \tblbest{0.616} & 0.287 & 0.628 & 0.823 & 0.510 & \tblsecondbest{0.505} & 0.660 & 0.890 \\
		
		\midrule
		
		 & \myalgonamefirst	& \tblsecondbest{0.426} & \tblsecondbest{0.488} & \tblsecondbest{0.244} & \tblsecondbest{0.481} & \tblsecondbest{0.609} & \tblsecondbest{0.343} & \tblsecondbest{0.277} & \tblsecondbest{0.452} & \tblsecondbest{0.617} \\
		& \myalgonamesecond	& \tblsecondbest{0.518} & \tblsecondbest{0.616} & \tblsecondbest{0.326}	& \tblsecondbest{0.599} & \tblsecondbest{0.768} & 0.475 & 0.452 & \tblbest{0.608}	& \tblbest{0.809} \\
		\multirow{-3}{*}{$\dfiltcoeff = 0.6$} & \myalgonamethird	& \tblsecondbest{0.507}	& \tblsecondbest{0.599}	& \tblbest{0.295}	& \tblbest{0.639} & \tblbest{0.827}	& \tblbest{0.514}	& \tblbest{0.510}	& \tblbest{0.683} & \tblbest{0.901} \\
		
		\midrule
		
		 & \myalgonamefirst	& 0.404 & 0.449 & 0.231 & 0.430 & 0.552 & 0.334 & 0.260 & 0.390 & 0.522 \\
		& \myalgonamesecond	& 0.513 & 0.603 & 0.310	& 0.594 & 0.766 & \tblsecondbest{0.478} & \tblsecondbest{0.456} & 0.586 & 0.781 \\
		\multirow{-3}{*}{$\dfiltcoeff = 0.7$} & \myalgonamethird	& \tblsecondbest{0.507}	& \tblsecondbest{0.599}	& \tblsecondbest{0.289}	& \tblsecondbest{0.638} & \tblbest{0.827} & \tblsecondbest{0.513} & \tblsecondbest{0.505} & \tblsecondbest{0.675} & \tblsecondbest{0.894} \\
		
		\midrule
		
		 & \myalgonamefirst	& 0.326 & 0.344 & 0.155 & 0.387 & 0.489 & 0.243 & 0.170 & 0.323 & 0.414 \\
		& \myalgonamesecond	& 0.505 & 0.598 & 0.297	& 0.592 & 0.764 & 0.457 & 0.426 & 0.589 & 0.774 \\
		\multirow{-3}{*}{$\dfiltcoeff = 0.8$} & \myalgonamethird	& 0.494	& 0.575	& 0.260 & 0.624 & 0.815 & 0.494 & 0.482	& 0.672 & 0.888 \\
		
		\midrule
		
		 & \myalgonamefirst	& 0.140 & 0.070 & 0.014 & 0.064 & 0.045 & 0.086 & 0.019 & 0.132 & 0.104 \\
		& \myalgonamesecond	& 0.471 & 0.541 & 0.250	& 0.547 & 0.697 & 0.445 & 0.409 & 0.574	& 0.746 \\
		\multirow{-3}{*}{$\dfiltcoeff = 0.9$} & \myalgonamethird	& 0.403	& 0.425	& 0.169 & 0.534 & 0.743 & 0.401 & 0.374 & 0.626 & 0.836 \\
		
		\bottomrule		
\end{tabular}
\end{table*}

%% file: tables/sota2.tex
\begin{table*}[t]
\fontsize{5}{6}\selectfont
	\centering
	\caption{Performance of the proposed trackers (in the last block of rows) in comparison with the the state-of-the-art. First block of rows reports the performance of the selected teachers; second block shows generic-approach tracker performance; third presents trackers that exploit experts or perform fusion. Best results are highlighted in red, second-best in blue.}
	\label{tab:sota}
	\setlength\tabcolsep{.08cm}
	\resizebox{\textwidth}{!}{
	\begin{tabular}{l | c c c |  c c |  c c | c c | c c c | c}
		\toprule
		
					& \multicolumn{3}{c|}{GOT-10k} & \multicolumn{2}{c|}{UAV123} & \multicolumn{2}{c|}{LaSOT} &  \multicolumn{2}{c|}{OTB-100} &  \multicolumn{3}{c|}{VOT2019} & \\
		\multirow{-2}{*}{Tracker} 	& AO & SR$_{0.50}$ & SR$_{0.75}$ & SS & PS  & SS  & PS & SS & PS & EAO & A & R & \multirow{-2}{*}{FPS} \\
		\midrule
		KCF \cite{KCF} & 0.203 & 0.177 & 0.065 & 0.331 & 0.503 & 0.178 & 0.166 & 0.477 &	0.693 & 0.110 & 0.441 & 1.279 & 105 \\
		MDNet \cite{MDNet} & 0.299& 0.303 & 0.099 & 0.489 & 0.718 & 0.397 & 0.373 & 0.673 & 0.909 & 0.151 & 0.507 & 0.782 & 5 \\
		ECO \cite{ECO} & 0.316 & 0.309	& 0.111	& 0.532	& 0.726	& 0.324	& 0.301	& 0.668	& 0.896 & 0.262 & 0.505 & 0.441 & 15 \\
		SiamRPN	\cite{SiamRPN} & 0.508	& 0.604	& 0.308	& 0.616	& 0.785	& 0.508	& 0.492	& 0.649 & 0.851 & 0.259 & 0.554 & 0.572 & 43 \\
		ATOM	\cite{ATOM} & 0.556	& 0.634	& 0.402	& 0.643	& 0.832 & 0.516	& 0.506	& 0.660 & 0.867 & \tblsecondbest{0.292} & \tblbest{0.603} & \tblsecondbest{0.411} & 20 \\
		DiMP	\cite{DiMP} & \tblsecondbest{0.611} & \tblsecondbest{0.717} & \tblbest{0.492}	& \tblsecondbest{0.653}	& \tblsecondbest{0.839}	& \tblsecondbest{0.570}	& \tblsecondbest{0.569} & 0.681 & 0.888 & \tblbest{0.379} & 0.594 & \tblbest{0.278} & 25 \\
		\midrule
		GOTURN	\cite{GOTURN} & 0.347	& 0.375	& 0.124	& 0.389	& 0.548	& 0.214	& 0.175	& 0.395 & 0.534 & - & - & - & 100 \\
		RE3	\cite{RE3} & -	& -	& -	& 0.514	& 0.667	& 0.325	& 0.301	& 0.464 & 0.582 & 0.152 & 0.458 & 0.940 & 150 \\
		ADNet	\cite{Yun2017} & -	& -	& -	& -	& -	& -	& -	& 0.646 & 0.880 & - & - & - & 3 \\
		ACT	\cite{Chen2018} & -	& -	& -	& 0.415	& 0.636	& -	& -	& 0.625 & 0.859 & - & - & - & 30 \\
		DRL-IS	\cite{Ren2018} & -	& -	& -	& -	& -	& -	& -	& 0.671 & 0.909 & - & - & - & 10 \\
		SiamRPN++	\cite{SiamRPNpp} & -	& -	& -	& 0.613	& 0.807	& 0.496	& -	& \tblsecondbest{0.696} & \tblsecondbest{0.914} & 0.285 & \tblsecondbest{0.599} & 0.482 & 35 \\
		GCT	\cite{GCT} & -	& -	& -	& 0.508	& 0.732	& -	& -	& 0.648 & 0.854 & - & - & - & 50 \\
		GradNet	\cite{GradNet} & -	& -	& -	& -	& -	& 0.365	& 0.351	& 0.639 & 0.861 & - & - & - & 80 \\
		SiamCAR	\cite{SiamCAR} & 0.569	& 0.670	& 0.415	& 0.614	& 0.760	& 0.507	& 0.510	& - & - & - & - & - & 52 \\
		ROAM	\cite{ROAM} & 0.436	& 0.466	& 0.164	& -	& -	& 0.368	& 0.390	& 0.681 & 0.908 & - & - & - & 13 \\
		\midrule
		MEEM	\cite{MEEM} & 0.253	& 0.235	& 0.068	& 0.392	& 0.627	& 0.280	& 0.224	& 0.566 & 0.830 & - & - & - & 10 \\
		HMMTxD	\cite{HMMTxD} & -	& -	& -	& -	& -	& -	& -	& - & - & 0.163 & 0.499 & 1.073 & - \\
		HDT	\cite{Qi2016} & -	& -	& -	& -	& -	& -	& -	& 0.562 & 0.844 & - & - & - & 10 \\
		Zhu et al. \cite{Zhu2018} & -	& -	& -	& -	& -	& -	& -	& 0.587 & 0.788 & - & - & - & 36 \\
		Li et al.	\cite{Li2019} & -	& -	& -	& -	& -	& -	& -	& 0.621 & 0.864 & - & - & - & 6 \\
		\midrule
		\myalgonamefirst			 & 0.484 & 0.556 & 0.326 & 0.515 & 0.655 & 0.386 & 0.330 & 0.481 & 0.644 & 0.131 & 0.400 & 1,020 & 90 \\
		\myalgonamesecond			 & 0.604 & 0.708 & 0.469	& 0.647 & 0.837 & 0.545 & 0.524 & 0.643	& 0.865 & 0.203 & 0.517 & 0.693  & 25 \\
		\myalgonamethird		& \tblbest{0.617}	& \tblbest{0.729} & \tblsecondbest{0.490}	& \tblbest{0.679} & \tblbest{0.873}	& \tblbest{0.576}	& \tblbest{0.574}	& \tblbest{0.701} & \tblbest{0.931} & 0.266 & 0.592 & 0.597 & 20 \\
		\bottomrule		
\end{tabular}
}
\end{table*}

%% file: 6conclusions.tex
\section{Conclusions}
In this paper, a novel methodology for visual tracking is proposed. KD and RL are joined in a novel framework where off-the-shelf tracking algorithms are employed to compress knowledge into a CNN-based student model. After learning, the student can be exploited in three different tracking setups, \myalgonamefirst, \myalgonamesecond and \myalgonamethird, depending on application needs. An extensive validation shows that the proposed trackers \myalgonamefirst and \myalgonamesecond compete with the state-of-the-art, while \myalgonamethird outperforms recently published methods and fusion approaches. All trackers can run in real-time.

%% file: supplementary.tex
\appendix

\begin{center}
Supplementary Material of \\
\Large
\textbf{Tracking-by-Trackers \\
with a Distilled and Reinforced Model}
\end{center}


\section{Methodology}

\subsection{MDP Auxiliary Functions}
\label{sec:mdpaux}

The function $\psi: \actions \times \reals^4 \xrightarrow{} \reals^4$ used to obtain the bounding box $\bbox_t$ given $\action_t$ and the previous bounding box $\bbox_{t-1}$ is defined such that
\begin{align}
    \psi(\action_t, \bbox_{t-1}) = 
    \begin{cases}
    x_t = x_{t-1} + \Delta x_t \cdot w_{t-1}\\
    y_t = y_{t-1} + \Delta y_t \cdot h_{t-1}\\
    w_t = w_{t-1} + \Delta w_t \cdot w_{t-1}\\
    h_t = h_{t-1} + \Delta h_t \cdot h_{t-1}\\
    \end{cases}
\end{align}.
The function $\phi : \reals^4 \times \reals^4 \xrightarrow{} \actions$ employed to obtain the expert action $\action^{(\teacher)}_t$ given the teacher bounding boxes $\bbox^{(\teacher)}_{t}, \bbox^{(\teacher)}_{t-1}$ is defined as
\begin{align}
\phi(\bbox^{(\teacher)}_{t}, \bbox^{(\teacher)}_{t-1}) = 
    \begin{cases}
    \Delta x^{(\teacher)}_t = (x^{(\teacher)}_{t} - x^{(\teacher)}_{t-1}) / w^{(\teacher)}_{t-1} \\
    \Delta y^{(\teacher)}_t = (y^{(\teacher)}_{t} - y^{(\teacher)}_{t-1}) / h^{(\teacher)}_{t-1} \\
    \Delta w^{(\teacher)}_t = (w^{(\teacher)}_{t} - w^{(\teacher)}_{t-1}) / w^{(\teacher)}_{t-1} \\
    \Delta h^{(\teacher)}_t = (h^{(\teacher)}_{t} - h^{(\teacher)}_{t-1}) / h^{(\teacher)}_{t-1} \\
    \end{cases}
\end{align}.

\subsection{Curriculum Learning Strategy}
\label{sec:curriculum}
A curriculum learning strategy \cite{Bengio2009} is designed to further facilitate and improve the student's learning. 
After terminating each $\episode_j$, a success counter $\counter_j$ for $\mdp_j$ is increased if $\student$ performs better than $\teacher$ in that interaction, i.e. if the first cumulative reward, received up to $\widehat{T}_j$, is greater or equal to the one obtained by the second. In formal terms, $\counter_j$ is updated if the following condition holds
\begin{align}
\sum_{i=1}^{\widehat{T}_j} \reward_i \geq \sum_{i=1}^{\widehat{T}_j} \reward^{(\teacher)}_i.
\end{align}
The counter update is done by testing students that interact with $\mdp_j$ by exploiting $\student_{\pi}(\state_t \:|\: \theta')$. 
The terminal video index $1 \geq \widehat{T}_j \geq T_j$ is successively increased during the training procedure by a central process which checks if
$\frac{\counter_j}{\episode_j} \geq \tau$. After each update of $\widehat{T}_j$, $\counter_j$ is reset to zero. 
By this setup, we ensure that, at every increase of $\widehat{T}_j$, students face a simpler learning problem where they are likely to succeed and in a shorter time, since they have already developed a tracking policy that, up to $\widehat{T}_j-1$, is at least good as the one of $\teacher$. We found $\tau = 0.25$ to work well in practice.


\section{Experimental Setup}

\subsubsection{Benchmarks and Performance Measures.}
\label{sec:benchmarks}
In this subsection we offer more details about the benchmark datasets and the relative performance measures employed to validate our methodology.

\paragraph{GOT-10k Test Set.}
The GOT-10k \cite{GOT10k} test set is composed of 180 videos. Target objects belong to 84 different classes, and 32 forms of object motion are present. An interesting note is that, except for the class \emph{person}, there is no overlap between the object classes in the training and test splits. For the \emph{person} class, there is no overlap in the type of motion. 
The evaluation protocol proposed by the authors is the one-pass evaluation (OPE) \cite{OTB}, while the metrics used are the average overlap (AO) and the success rates (SR) with overlap thresholds $0.50$ and $0.75$. 

\paragraph{OTB-100.}
\label{sec:otb100res}
The OTB-100 \cite{OTB} benchmark is a set of 100 challenging videos and it is widely used in the tracking literature. The standard evaluation procedure for this dataset is the OPE method while the Area Under the Curve (AUC) of the success and precision plot, referred as success score (SS) and precision scores (PS) respectively, are utilized to quantify trackers' performance. 

\paragraph{UAV123.}
The UAV123  benchmark \cite{UAV123} proposes 123 videos that are inherently different from traditional visual tracking benchmarks like OTB and VOT, since it offers sequences acquired form low-altitude UAVs. 
To evaluate trackers, the standard OTB methodology \cite{OTB} is exploited.  
 
\paragraph{LaSOT.}
A performance evaluation was also performed on the test set of LaSOT benchmark \cite{LaSOT}. This dataset is composed of 280 videos, with a total of more than 650k frames and an average sequence length of 2500 frames, that is higher than the lengths of the videos contained in the aforementioned benchmarks. The same methodology and metrics used for the OTB \cite{OTB} experiments are employed.
 
\paragraph{VOT2019.}
The VOT benchmarks are datasets used in the annual VOT tracking competitions. These sets change year by year, introducing increasingly challenging tracking scenarios. 
We evaluated our trackers on the set of the VOT2019 challenge \cite{VOT2019}, which provides 60 highly challenging videos.
Within the framework used by the VOT committee, trackers are evaluated based on Expected Average Overlap (EAO), Accuracy (A) and Robustness (R)~\cite{VOT}. Differently from the OPE, the VOT evaluation protocol presents the automatic re-initialization of the tracker when the IoU between its estimated bounding box and the ground-truth becomes zero.


\section{Additional Results}

\subsection{Impact of Transfer Set}
We evaluated how performance change considering other sources of video data. By respecting the idea that unbiased demonstrations of the teachers should be employed, we used the training set of the LaSOT benchmark \cite{LaSOT}. This dataset is smaller than the training set of GOT-10k and contains 1120 videos with approximately 2.83M frames. After filtering the trajectories, we obtained the transfer set $\dataset$ which specification are given in Table \ref{tab:demostatslasot}.

\input{tables/demostatslasot.tex}

The results are shown in Table \ref{tab:demofiltlasot}. The amount of training samples is lower than the amount obtained by filtering the GOT-10k transfer set with $\dfiltcoeff = 0.8$, and the proposed trackers present a behaviour that reflects the loss of data (as seen in Table \ref{tab:demofilt}). This experiment suggests that the quantity of data has more impact than the quality of data.
\input{tables/demofiltlasot.tex}

\subsection{Success and Precision Plots on OTB-100}
\begin{figure}[!htbp]
\begin{center}
\includegraphics[width=.9\columnwidth]{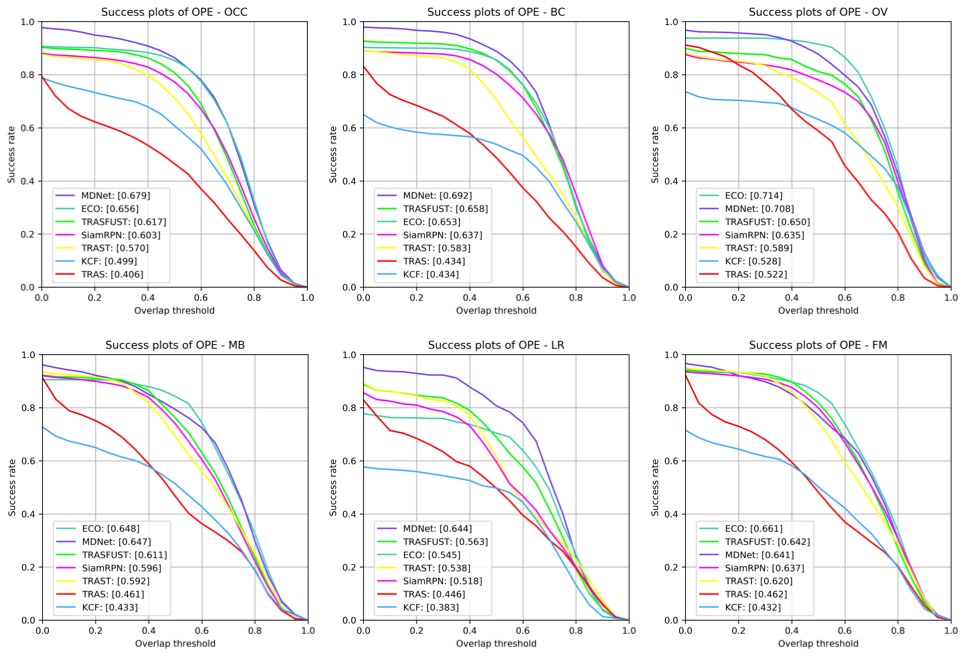}
\end{center}
   \caption{Success plots on OTB-100 presenting the performance of the proposed trackers and the teachers on tracking situations with: occlusion (OCC); background clutter (BC); out of view (OV); motion blur (MB); low resolution (LR); fast motion (FM).}
\label{fig:otb2015succ}
\end{figure}

\begin{figure}[!htbp]
\begin{center}
\includegraphics[width=.9\columnwidth]{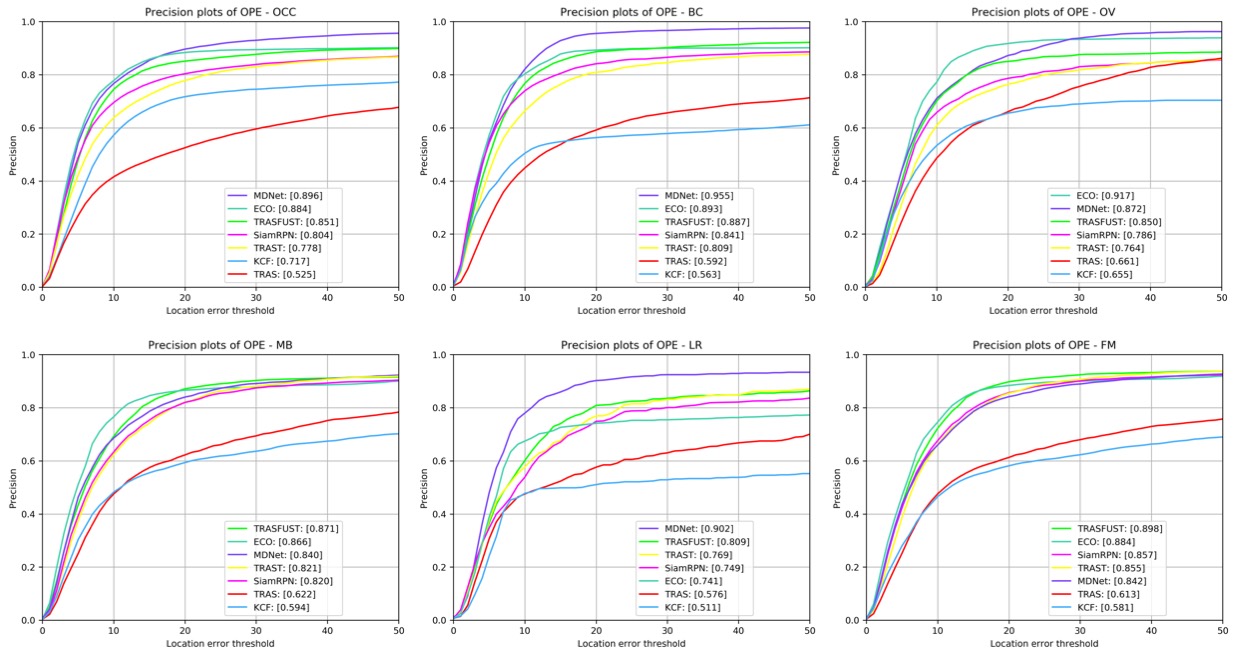}
\end{center}
   \caption{Precision plots on OTB-100 presenting the performance of the proposed trackers and the teachers on tracking situations with: occlusion (OCC); background clutter (BC); out of view (OV); motion blur (MB); low resolution (LR); fast motion (FM).}
\label{fig:otb2015prec}
\end{figure}

In Figures \ref{fig:otb2015succ} and \ref{fig:otb2015prec} the success plots and precision plots for different sequence categories of the OTB-100 benchmark are presented. 

\subsection{Video}
At this link \url{https://youtu.be/uKtQgPk3nCU}, we provide a video showing the tracking abilities of our proposed trackers. For each video, the predictions of \myalgonamefirst, \myalgonamesecond and \myalgonamethird are shown. For \myalgonamesecond and \myalgonamethird, we report also the tracker which prediction was chosen as output proposes (with the term "CONTROLLING" next to the tracker's name).

%% file: tables/demostatslasot.tex
\begin{table*}[t]
\fontsize{5}{6}\selectfont
	\centering
	\caption{Teacher-based statistics of the LaSOT transfer set.}
	\label{tab:demostatslasot}
	\setlength\tabcolsep{.05cm}
	\begin{tabular}{l | c c c }
		\toprule
		Teachers & \multicolumn{3}{c}{$\dfiltcoeff = 0.5$} \\
		 & \# traj & AO & $|\mathcal{D}|$ \\
		\midrule
		$\tk$ & 16 & 0.835 & 80 \\
		$\tm$ & 32 & 0.830 & 160 \\
		$\te$ & 44 & 0.817 & 220 \\
		$\ts$ & 87 & 0.852 & 435 \\
		$\tp$ & 106 & 0.856	& 530 \\
		\bottomrule		
\end{tabular}
\end{table*}

%% file: tables/demofiltlasot.tex
\begin{table*}[t]
\fontsize{5}{6}\selectfont
	\centering
	\caption{Performance of the proposed trackers considering the training set of LaSOT as transfer set.}
	\label{tab:demofiltlasot}
	\setlength\tabcolsep{.08cm}
	\begin{tabular}{l | l | c c c | c c | c c | c c }
		\toprule
		& & \multicolumn{3}{c|}{GOT-10k} & \multicolumn{2}{c|}{UAV123} & \multicolumn{2}{c|}{LaSOT} &  \multicolumn{2}{c}{OTB-100}  \\
		
		& \multirow{-2}{*}{Tracker} 	& AO & SR$_{0.50}$ & SR$_{0.75}$ & SS & PS  & SS  & PS & SS & PS \\
		\midrule
		
		 & \myalgonamefirst	& 0.242 & 0.252 & 0.086 & 0.329 & 0.437 & 0.222 & 0.166 & 0.254 & 0.337 \\
		 & \myalgonamesecond	& 0.475 & 0.552 & 0.248	&  0.553 & 0.746 & 0.463 & 0.432 & 0.577 & 0.760 \\
		 \multirow{-3}{*}{$\dfiltcoeff = 0.5$} & \myalgonamethird	& 0.468	& 0.529 & 0.221 & 0.594 & 0.803 & 0.470 & 0.452 & 0.666 & 0.885 \\
		
		\bottomrule		
\end{tabular}
\end{table*}

%% file: accv2020cameraready.bbl
\begin{thebibliography}{10}

\bibitem{Bolme2010}
Bolme, D.S., Beveridge, J.R., Draper, B.A., Lui, Y.M.:
\newblock {Visual object tracking using adaptive correlation filters}.
\newblock In: IEEE Conference on Computer Vision and Pattern Recognition, IEEE
  (2010)  2544--2550

\bibitem{KCF}
Henriques, J.F., Caseiro, R., Martins, P., Batista, J.:
\newblock {High-speed tracking with kernelized correlation filters}.
\newblock IEEE Transactions on Pattern Analysis and Machine Intelligence
  \textbf{37} (2015)  583--596

\bibitem{DSST}
Danelljan, M., Hager, G., Khan, F.S., Felsberg, M.:
\newblock {Discriminative Scale Space Tracking}.
\newblock IEEE Transactions on Pattern Analysis and Machine Intelligence
  \textbf{39} (2017)  1561--1575

\bibitem{Staple}
Bertinetto, L., Valmadre, J., Golodetz, S., Miksik, O., Torr, P.H.:
\newblock {Staple: Complementary learners for real-time tracking}.
\newblock In: IEEE Conference on Computer Vision and Pattern Recognition.
  Volume 2016-Decem. (2016)  1401--1409

\bibitem{Lukezic2018}
Luke{\v{z}}i{\v{c}}, A., Voj{\'{i}}ř, T., {{\v{C}}ehovin Zajc}, L., Matas, J.,
  Kristan, M.:
\newblock {Discriminative Correlation Filter Tracker with Channel and Spatial
  Reliability}.
\newblock International Journal of Computer Vision \textbf{126} (2018)
  671--688

\bibitem{GOTURN}
Held, D., Thrun, S., Savarese, S.:
\newblock {Learning to Track at 100 FPS with Deep Regression Networks}.
\newblock In: European Conference on Computer Vision. Volume abs/1604.0. (2016)

\bibitem{RE3}
Gordon, D., Farhadi, A., Fox, D.:
\newblock {Re 3 : Real-time recurrent regression networks for visual tracking
  of generic objects}.
\newblock IEEE Robotics and Automation Letters \textbf{3} (2018)  788--795

\bibitem{SiamFC}
Bertinetto, L., Valmadre, J., Henriques, J.F., Vedaldi, A., Torr, P.H.:
\newblock {Fully-convolutional siamese networks for object tracking}.
\newblock European Conference on Computer Vision \textbf{9914 LNCS} (2016)
  850--865

\bibitem{SiamRPN}
Li, B., Yan, J., Wu, W., Zhu, Z., Hu, X.:
\newblock {High Performance Visual Tracking with Siamese Region Proposal
  Network}.
\newblock In: IEEE Conference on Computer Vision and Pattern Recognition, IEEE
  (2018)  8971--8980

\bibitem{SiamRPNpp}
Li, B., Wu, W., Wang, Q., Zhang, F., Xing, J., Yan, J.:
\newblock {SIAMRPN++: Evolution of siamese visual tracking with very deep
  networks}.
\newblock IEEE Conference on Computer Vision and Pattern Recognition
  \textbf{2019-June} (2019)  4277--4286

\bibitem{DaSiam}
Zhu, Z., Wang, Q., Li, B., Wu, W., Yan, J., Hu, W.:
\newblock {Distractor-aware siamese networks for visual object tracking}.
\newblock In: European Conference on Computer Vision. Volume 11213 LNCS. (2018)
   103--119

\bibitem{Zhang2019}
Zhang, Z., Peng, H.:
\newblock {Deeper and Wider Siamese Networks for Real-Time Visual Tracking}.
\newblock IEEE Conference on Computer Vision and Pattern Recognition (2019)

\bibitem{MDNet}
Nam, H., Han, B.:
\newblock {Learning Multi-domain Convolutional Neural Networks for Visual
  Tracking}.
\newblock IEEE Conference on Computer Vision and Pattern Recognition
  \textbf{2016-Decem} (2016)  4293--4302

\bibitem{RealTimeMDNet}
Jung, I., Son, J., Baek, M., Han, B.:
\newblock {Real-Time MDNet}.
\newblock In: European Conference on Computer Vision. (2018)

\bibitem{ECO}
Danelljan, M., Bhat, G., Khan, F.S., Felsberg, M.:
\newblock {ECO: Efficient Convolution Operators for Tracking}.
\newblock In: IEEE Conference on Computer Vision and Pattern Recognition.
  (2017)

\bibitem{ATOM}
Danelljan, M., Bhat, G., Khan, F.S., Felsberg, M.:
\newblock {ATOM: Accurate Tracking by Overlap Maximization}.
\newblock In: IEEE Conference on Computer Vision and Pattern Recognition.
  (2019)

\bibitem{DiMP}
Bhat, G., Danelljan, M., {Van Gool}, L., Timofte, R.:
\newblock {Learning Discriminative Model Prediction for Tracking}.
\newblock In: Proceedings of the IEEE/CVF International Conference on Computer
  Vision. (2019)

\bibitem{MEEM}
Zhang, J., Ma, S., Sclaroff, S.:
\newblock {MEEM: Robust tracking via multiple experts using entropy
  minimization}.
\newblock In: European Conference on Computer Vision. Volume 8694 LNCS.,
  Springer Verlag (2014)  188--203

\bibitem{Yoon2012}
Yoon, J.H., Kim, D.Y., Yoon, K.J.:
\newblock {Visual tracking via adaptive tracker selection with multiple
  features}.
\newblock In: European Conference on Computer Vision. Volume 7575 LNCS. (2012)
  28--41

\bibitem{Wang2014}
Wang, N., Yeung, D.Y.:
\newblock {Ensemble-based tracking: Aggregating crowdsourced structured time
  series data}.
\newblock In: 31st International Conference on Machine Learning, ICML 2014.
  Volume~4. (2014)  2807--2817

\bibitem{Bailer2014}
Bailer, C., Pagani, A., Stricker, D.:
\newblock {A superior tracking approach: Building a strong tracker through
  fusion}.
\newblock In: European Conference on Computer Vision. Volume 8695 LNCS.,
  Springer Verlag (2014)  170--185

\bibitem{HMMTxD}
Vojir, T., Matas, J., Noskova, J.:
\newblock {Online adaptive hidden Markov model for multi-tracker fusion}.
\newblock Computer Vision and Image Understanding \textbf{153} (2016)  109--119

\bibitem{Comanciu2000}
Comaniciu, D., Ramesh, V., Meer, P.:
\newblock {Real-time tracking of non-rigid objects using mean shift}.
\newblock IEEE Conference on Computer Vision and Pattern Recognition \textbf{2}
  (2000)  142--149

\bibitem{Matrioska}
Maresca, M.E., Petrosino, A.:
\newblock {MATRIOSKA: A multi-level approach to fast tracking by learning}.
\newblock In: International Conference on Image Analysis and Processing. Volume
  8157 LNCS. (2013)  419--428

\bibitem{LGT}
{\v{C}}ehovin, L., Kristan, M., Leonardis, A.:
\newblock {Robust visual tracking using an adaptive coupled-layer visual
  model}.
\newblock IEEE Transactions on Pattern Analysis and Machine Intelligence
  \textbf{35} (2013)  941--953

\bibitem{OGT}
Nam, H., Hong, S., Han, B.:
\newblock {Online graph-based tracking}.
\newblock In: European Conference on Computer Vision. Volume 8693 LNCS.,
  Springer Verlag (2014)  112--126

\bibitem{Struck}
Hare, S., Golodetz, S., Saffari, A., Vineet, V., Cheng, M.M., Hicks, S.L.,
  Torr, P.H.:
\newblock {Struck: Structured Output Tracking with Kernels}.
\newblock IEEE Transactions on Pattern Analysis and Machine Intelligence
  \textbf{38} (2016)  2096--2109

\bibitem{Yun2017}
Yun, S., Choi, J., Yoo, Y., Yun, K., Choi, J.Y.:
\newblock {Action-decision networks for visual tracking with deep reinforcement
  learning}.
\newblock In: IEEE Conference on Computer Vision and Pattern Recognition.
  Volume 2017-Janua., IEEE (2017)  1349--1358

\bibitem{Supancic2017}
Supancic, J., Ramanan, D.:
\newblock {Tracking as Online Decision-Making: Learning a Policy from Streaming
  Videos with Reinforcement Learning}.
\newblock Proceedings of the IEEE International Conference on Computer Vision
  \textbf{2017-Octob} (2017)  322--331

\bibitem{Choi2017}
Choi, J., Kwon, J., Lee, K.M.:
\newblock {Real-time visual tracking by deep reinforced decision making}.
\newblock Computer Vision and Image Understanding \textbf{171} (2018)  10--19

\bibitem{Ren2018}
Ren, L., Yuan, X., Lu, J., Yang, M., Zhou, J.:
\newblock {Deep Reinforcement Learning with Iterative Shift for Visual
  Tracking}.
\newblock In: European Conference on Computer Vision. (2018)  684--700

\bibitem{Chen2018}
Chen, B., Wang, D., Li, P., Wang, S., Lu, H.:
\newblock {Real-time 'Actor-Critic' Tracking}.
\newblock In: European Conference on Computer Vision. (2018)  318--334

\bibitem{Dunnhofer2019}
Dunnhofer, M., Martinel, N., Foresti, G.L., Micheloni, C.:
\newblock {Visual Tracking by means of Deep Reinforcement Learning and an
  Expert Demonstrator}.
\newblock In: Proceedings of The IEEE/CVF International Conference on Computer
  Vision Workshops. (2019)

\bibitem{CCOT}
Danelljan, M., Robinson, A., Khan, F.S., Felsberg, M.:
\newblock {Beyond Correlation Filters: Learning Continuous Convolution
  Operators for Visual Tracking}.
\newblock In: European Conference on Computer Vision. (2016)

\bibitem{SiamMask}
Wang, Q., Zhang, L., Bertinetto, L., Hu, W., Torr, P.H.S.:
\newblock {Fast Online Object Tracking and Segmentation: A Unifying Approach}.
\newblock In: IEEE Conference on Computer Vision and Pattern Recognition.
  (2019)

\bibitem{Dunnhofer2020MedIA}
Dunnhofer, M., Antico, M., Sasazawa, F., Takeda, Y., Camps, S., Martinel, N.,
  Micheloni, C., Carneiro, G., Fontanarosa, D.:
\newblock {Siam-U-Net: encoder-decoder siamese network for knee cartilage
  tracking in ultrasound images}.
\newblock Medical Image Analysis (2020)

\bibitem{Hinton2014KD}
Hinton, G., Vinyals, O., Dean, J.:
\newblock {Distilling the Knowledge in a Neural Network}.
\newblock In: Deep Learning Workshop NIPS 2014. (2014)

\bibitem{He2016ResNet}
He, K., Zhang, X., Ren, S., Sun, J.:
\newblock {Deep residual learning for image recognition}.
\newblock In: IEEE Conference on Computer Vision and Pattern Recognition.
  Volume 2016-Decem. (2016)  770--778

\bibitem{Tang2016}
Tang, Z., Wang, D., Zhang, Z.:
\newblock {Recurrent neural network training with dark knowledge transfer}.
\newblock In: IEEE International Conference on Acoustics, Speech and Signal
  Processing. Volume 2016-May. (2016)  5900--5904

\bibitem{Li2017KDnoise}
Li, Y., Yang, J., Song, Y., Cao, L., Luo, J., Li, L.J.:
\newblock {Learning from Noisy Labels with Distillation}.
\newblock In: Proceedings of the IEEE International Conference on Computer
  Vision. Volume 2017-Octob. (2017)  1928--1936

\bibitem{Phuong2019}
Phuong, M., Lampert, C.H.:
\newblock {Distillation-Based Training for Multi-Exit Architectures}.
\newblock In: Proceedings of the IEEE/CVF International Conference on Computer
  Vision. (2019)

\bibitem{Geras2015}
Geras, K.J., Mohamed, A.r., Caruana, R., Urban, G., Wang, S., Aslan, O.,
  Philipose, M., Richardson, M., Sutton, C.:
\newblock {Blending LSTMs into CNNs}.
\newblock (2015)

\bibitem{Chen2017}
Chen, G., Choi, W., Yu, X., Han, T., Chandraker, M.:
\newblock {Learning efficient object detection models with knowledge
  distillation}.
\newblock In: Advances in Neural Information Processing Systems. Volume
  2017-Decem. (2017)  743--752

\bibitem{Howard2017}
Howard, A.G., Zhu, M., Chen, B., Kalenichenko, D., Wang, W., Weyand, T.,
  Andreetto, M., Adam, H.:
\newblock {MobileNets: Efficient Convolutional Neural Networks for Mobile
  Vision Applications}.
\newblock (2017)

\bibitem{Polino2018}
Polino, A., Pascanu, R., Alistarh, D.:
\newblock {Model compression via distillation and quantization}.
\newblock In: International Conference on Learning Representations,
  International Conference on Learning Representations, ICLR (2018)

\bibitem{Watkins1992}
Watkins, C.J.C.H., Dayan, P.:
\newblock {Q-learning}.
\newblock Machine Learning \textbf{8} (1992)  279--292

\bibitem{Konda2000}
Konda, V.R., Tsitsiklis, J.N.:
\newblock {Actor-Critic Algorithms}.
\newblock In: Advances in Neural Information Processing Systems. (2000)

\bibitem{Sutton2000}
Sutton, R.S., McAllester, D., Singh, S., Mansour, Y.:
\newblock {Policy gradient methods for reinforcement learning with function
  approximation}.
\newblock In: Advances in Neural Information Processing Systems. (2000)
  1057--1063

\bibitem{Mnih2013}
Mnih, V., Kavukcuoglu, K., Silver, D., Graves, A., Antonoglou, I., Wierstra,
  D., Riedmiller, M.:
\newblock {Playing Atari with Deep Reinforcement Learning}.
\newblock CoRR \textbf{abs/1312.5} (2013)

\bibitem{Mnih2016}
Mnih, V., Badia, A.P., Mirza, L., Graves, A., Harley, T., Lillicrap, T.P.,
  Silver, D., Kavukcuoglu, K.:
\newblock {Asynchronous methods for deep reinforcement learning}.
\newblock 33rd International Conference on Machine Learning, ICML 2016
  \textbf{4} (2016)  2850--2869

\bibitem{TLD}
Kalal, Z., Mikolajczyk, K., Matas, J.:
\newblock {Tracking-learning-detection}.
\newblock IEEE Transactions on Pattern Analysis and Machine Intelligence
  \textbf{34} (2012)  1409--1422

\bibitem{Qi2016}
Qi, Y., Zhang, S., Qin, L., Yao, H., Huang, Q., Lim, J., Yang, M.H.:
\newblock {Hedged Deep Tracking}.
\newblock In: IEEE Conference on Computer Vision and Pattern Recognition.
  Volume 2016-Decem. (2016)  4303--4311

\bibitem{Li2019}
Li, Z., Wei, W., Zhang, T., Wang, M., Hou, S., Peng, X.:
\newblock {Online Multi-Expert Learning for Visual Tracking}.
\newblock IEEE Transactions on Image Processing \textbf{29} (2019)  934--946

\bibitem{Bucila2006}
Bucilǎ, C., Caruana, R., Niculescu-Mizil, A.:
\newblock {Model compression}.
\newblock In: Proceedings of the ACM SIGKDD International Conference on
  Knowledge Discovery and Data Mining. Volume 2006. (2006)  535--541

\bibitem{Rusu2016}
Rusu, A.A., Colmenarejo, S.G., G{\"{u}}l{\c{c}}ehre, {\c{C}}., Desjardins, G.,
  Kirkpatrick, J., Pascanu, R., Mnih, V., Kavukcuoglu, K., Hadsell, R.:
\newblock {Policy distillation}.
\newblock In: 4th International Conference on Learning Representations, ICLR
  2016. (2016)

\bibitem{Parisotto2016}
Parisotto, E., Ba, J., Salakhutdinov, R.:
\newblock {Actor-mimic deep multitask and transfer reinforcement learning}.
\newblock In: 4th International Conference on Learning Representations, ICLR
  2016, International Conference on Learning Representations, ICLR (2016)

\bibitem{Garcia2018KDAR}
Garcia, N.C., Morerio, P., Murino, V.:
\newblock {Modality distillation with multiple stream networks for action
  recognition}.
\newblock In: European Conference on Computer Vision. Volume 11212 LNCS. (2018)
   106--121

\bibitem{Wank2019KDAR}
Wang, X., Hu, J.F., Lai, J., Zhang, J., Zheng, W.S.:
\newblock {Progressive Teacher-student Learning for Early Action Prediction}.
\newblock Computer Vision and Pattern Recognition (CVPR) (2019)  3556--3565

\bibitem{Shmelkov2017}
Shmelkov, K., Schmid, C., Alahari, K.:
\newblock {Incremental Learning of Object Detectors without Catastrophic
  Forgetting}.
\newblock In: Proceedings of the IEEE International Conference on Computer
  Vision. Volume 2017-Octob. (2017)  3420--3429

\bibitem{Liu2019semantic}
Liu, Y., Chen, K., Liu, C., Qin, Z., Luo, Z., Wang, J.:
\newblock {Structured Knowledge Distillation for Semantic Segmentation}.
\newblock In: IEEE Conference on Computer Vision and Pattern Recognition.
  (2019)  2599--2608

\bibitem{He2019}
He, T., Shen, C., Tian, Z., Gong, D., Sun, C., Yan, Y.:
\newblock {Knowledge Adaptation for Efficient Semantic Segmentation}.
\newblock In: IEEE Conference on Computer Vision and Pattern Recognition.
  (2019)  578--587

\bibitem{Wu2019KDreid}
Wu, A., Zheng, W.S., Guo, X., Lai, J.H.:
\newblock {Distilled Person Re-identification:Towards a More Scalable System}.
\newblock IEEE Conference on Computer Vision and Pattern Recognition (2019)
  1187--1196

\bibitem{Wang2019}
Wang, N., Zhou, W., Song, Y., Ma, C., Li, H.:
\newblock {Real-Time Correlation Tracking Via Joint Model Compression and
  Transfer}.
\newblock IEEE Transactions on Image Processing \textbf{29} (2020)  6123--6135

\bibitem{Liu2019}
Liu, Y., Dong, X., Lu, X., Khan, F.S., Shen, J., Hoi, S.:
\newblock {Teacher-Students Knowledge Distillation for Siamese Trackers}.
\newblock (2019)

\bibitem{Meshgi2019}
{Meshgi}, K., {Mirzaei}, M.S., {Oba}, S.:
\newblock Long and short memory balancing in visual co-tracking using
  q-learning.
\newblock In: 2019 IEEE International Conference on Image Processing (ICIP).
  (2019)  3970--3974

\bibitem{Gorila}
Nair, A., Srinivasan, P., Blackwell, S., Alcicek, C., Fearon, R., {De Maria},
  A., Panneershelvam, V., Suleyman, M., Beattie, C., Petersen, S., Legg, S.,
  Mnih, V., Kavukcuoglu, K., Silver, D.:
\newblock {Massively Parallel Methods for Deep Reinforcement Learning}.
\newblock (2015)

\bibitem{IMPALA}
Espeholt, L., Soyer, H., Munos, R., Simonyan, K., Mnih, V., Ward, T., Yotam,
  B., Vlad, F., Tim, H., Dunning, I., Legg, S., Kavukcuoglu, K.:
\newblock {IMPALA: Scalable Distributed Deep-RL with Importance Weighted
  Actor-Learner Architectures}.
\newblock In: 35th International Conference on Machine Learning, ICML 2018.
  Volume~4. (2018)  2263--2284

\bibitem{SuttonBarto2018}
Sutton, R.S., Barto, A.G.:
\newblock {Reinforcement Learning: An Introduction}. 2nd edn.
\newblock MIT Press, Cambridge, MA, USA (2018)

\bibitem{Bengio2009}
Bengio, Y., Louradour, J., Collobert, R., Weston, J.:
\newblock {Curriculum learning}.
\newblock In: 26th International Conference on Machine Learning, ICML '09, New
  York, New York, USA, ACM Press (2009)  1--8

\bibitem{Hochreiter1997LSTM}
Hochreiter, S., Schmidhuber, J.:
\newblock {Long Short-Term Memory}.
\newblock Neural Computation \textbf{9} (1997)  1735--1780

\bibitem{GOT10k}
Huang, L., Zhao, X., Huang, K.:
\newblock {GOT-10k: A Large High-Diversity Benchmark for Generic Object
  Tracking in the Wild}.
\newblock IEEE Transactions on Pattern Analysis and Machine Intelligence (2019)
   1--1

\bibitem{UAV123}
Mueller, M., Smith, N., Ghanem, B.:
\newblock {A Benchmark and Simulator for UAV Tracking}.
\newblock In: European Conference on Computer Vision, Springer, Cham (2016)
  445--461

\bibitem{LaSOT}
Fan, H., Lin, L., Yang, F., Chu, P., Deng, G., Yu, S., Bai, H., Xu, Y., Liao,
  C., Ling, H.:
\newblock {LaSOT: A High-quality Benchmark for Large-scale Single Object
  Tracking}.
\newblock In: IEEE Conference on Computer Vision and Pattern Recognition.
  (2019)

\bibitem{OTB}
Wu, Y., Lim, J., Yang, M.H.:
\newblock {Online object tracking: A benchmark}.
\newblock In: IEEE Conference on Computer Vision and Pattern Recognition, IEEE
  Computer Society (2013)  2411--2418

\bibitem{VOT2019}
Kristan, M., Matas, J., Leonardis, A., Felsberg, M., Pflugfelder, R.,
  K{\"{a}}m{\"{a}}r{\"{a}}inen, J.K., Zajc, L., Drbohlav, O.,
  Luke{\v{z}}i{\v{c}}, A., Berg, A., Eldesokey, A., K{\"{a}}pyl{\"{a}}, J.,
  Fern{\'{a}}ndez, G., Gonzalez-Garcia, A., Memarmoghadam, A., Lu, A., He, A.,
  Varfolomieiev, A., Chan, A., {Shekhar Tripathi}, A., Smeulders, A., {Suraj
  Pedasingu}, B., {Xin Chen}, B., Zhang, B., Wu, B., Li, B., He, B., Yan, B.,
  Bai, B., Li, B., Li, B., {Hak Kim}, B., Ma, C., Fang, C., Qian, C., Chen, C.,
  Li, C., Zhang, C., Tsai, C.Y., Luo, C., Micheloni, C., Zhang, C., Tao, D.,
  Gupta, D., Song, D., Wang, D., Gavves, E., Yi, E., {Shahbaz Khan}, F., Zhang,
  F., Wang, F., Zhao, F., {De Ath}, G., Bhat, G., Chen, G., Wang, G., Li, G.,
  Cevikalp, H., Du, H., Zhao, H., Saribas, H., {Min Jung}, H., Bai, H., Yu, H.,
  Peng, H., Lu, H., Li, H., Li, J., Li, J., Fu, J., Chen, J., Gao, J., Zhao,
  J., Tang, J., Li, J., Wu, J., Liu, J., Wang, J., Qi, J., Zhang, J., Tsotsos,
  J.K., {Hyuk Lee}, J., van~de Weijer, J., Kittler, J., {Ha Lee}, J., Zhuang,
  J., Zhang, K., Wang, K., Dai, K., Chen, L., Liu, L., Guo, L., Zhang, L.,
  Wang, L., Wang, L., Zhang, L., Wang, L., Zhou, L., Zheng, L., Rout, L., {Van
  Gool}, L., Bertinetto, L., Danelljan, M., Dunnhofer, M., Ni, M., {Young Kim},
  M., Tang, M., Yang, M.H., Paluru, N., Martinel, N., Xu, P., Zhang, P., Zheng,
  P., Zhang, P., Torr, P.H., {Zhang Qiang Wang}, Q., Guo, Q., Timofte, R.,
  {Krishna Gorthi}, R., Everson, R., Han, R., Zhang, R., You, S., Zhao, S.C.,
  Zhao, S., Li, S., Li, S., Ge, S., Bai, S., Guan, S., Xing, T., Xu, T., Yang,
  T., Zhang, T., Oj{\'{i}}, T., Feng, W., Hu, W., Wang, W., Tang, W., Zeng, W.,
  Liu, W., Chen, X., Qiu, X., Bai, X., Wu, X.J., Yang, X., Chen, X., Li, X.,
  Sun, X., Chen, X., Tian, X., Tang, X., Zhu, X.F., Huang, Y., Chen, Y., Lian,
  Y., Gu, Y., Liu, Y., Chen, Y., Zhang, Y., Xu, Y., Wang, Y., Li, Y., Zhou, Y.,
  Dong, Y., Xu, Y., Zhang, Y., Li, Y., {Wang Zhao Luo}, Z., Zhang, Z., Feng,
  Z.H., He, Z., Song, Z., Chen, Z., Zhang, Z., Wu, Z., Xiong, Z., Huang, Z.,
  Teng, Z., Ni, Z.:
\newblock {The Seventh Visual Object Tracking VOT2019 Challenge Results}.
\newblock In: Proceedings of the IEEE/CVF International Conference on Computer
  Vision Workshops. (2019)

\bibitem{VOT}
Kristan, M., Matas, J., Leonardis, A., Vojir, T., Pflugfelder, R., Fernandez,
  G., Nebehay, G., Porikli, F., {\v{C}}ehovin, L.:
\newblock {A Novel Performance Evaluation Methodology for Single-Target
  Trackers}.
\newblock IEEE Transactions on Pattern Analysis and Machine Intelligence
  \textbf{38} (2016)  2137--2155

\bibitem{ImageNet}
Deng, J., Dong, W., Socher, R., Li, L.J., {Kai Li}, {Li Fei-Fei}:
\newblock {ImageNet: A large-scale hierarchical image database}.
\newblock In: IEEE Conference on Computer Vision and Pattern Recognition, IEEE
  (2009)  248--255

\bibitem{Radam}
Liu, L., Jiang, H., He, P., Chen, W., Liu, X., Gao, J., Han, J.:
\newblock {On the Variance of the Adaptive Learning Rate and Beyond}.
\newblock (2019)

\bibitem{Cho2019}
Cho, J.H., Hariharan, B.:
\newblock {On the efficacy of knowledge distillation}.
\newblock In: Proceedings of the IEEE International Conference on Computer
  Vision. Volume 2019-Octob., Institute of Electrical and Electronics Engineers
  Inc. (2019)  4793--4801

\bibitem{GCT}
Gao, J., Zhang, T., Xu, C.:
\newblock {Graph Convolutional Tracking}.
\newblock In: IEEE Conference on Computer Vision and Pattern Recognition.
  Number~1 (2019)  4649--4659

\bibitem{GradNet}
Li, P., Chen, B., Ouyang, W., Wang, D., Yang, X., Lu, H.:
\newblock {GradNet: Gradient-Guided Network for Visual Object Tracking}.
\newblock In: Proceedings of the IEEE/CVF International Conference on Computer
  Vision. (2019)

\bibitem{SiamCAR}
Guo, D., Wang, J., Cui, Y., Wang, Z., Chen, S.:
\newblock {SiamCAR: Siamese Fully Convolutional Classification and Regression
  for Visual Tracking}.
\newblock In: IEEE/CVF Conference on Computer Vision and Pattern Recognition.
  (2020)

\bibitem{ROAM}
Yang, T., Xu, P., Hu, R., Chai, H., Chan, A.B.:
\newblock {ROAM: Recurrently Optimizing Tracking Model}.
\newblock In: IEEE/CVF Conference on Computer Vision and Pattern Recognition.
  (2020)

\bibitem{Zhu2018}
Zhu, Y., Wen, J., Zhang, L., Wang, Y.:
\newblock {Visual Tracking with Dynamic Model Update and Results Fusion}.
\newblock In: Proceedings - International Conference on Image Processing, IEEE
  Computer Society (2018)  2685--2689

\end{thebibliography}
